# Learning Symbolic Models of Stochastic Domains

**Hanna M. Pasula**                                   PASULA@CSAIL.MIT.EDU
**Luke S. Zettlemoyer**                                    LSZ@CSAIL.MIT.EDU
**Leslie Pack Kaelbling**                                 LPK@CSAIL.MIT.EDU
*MIT CSAIL*
*Cambridge, MA 02139*

## Abstract

In this article, we work towards the goal of developing agents that can learn to act in complex worlds. We develop a probabilistic, relational planning rule representation that compactly models noisy, nondeterministic action effects, and show how such rules can be effectively learned. Through experiments in simple planning domains and a 3D simulated blocks world with realistic physics, we demonstrate that this learning algorithm allows agents to effectively model world dynamics.

## 1. Introduction

One of the goals of artificial intelligence is to build systems that can act in complex environments as effectively as humans do: to perform everyday human tasks, like making breakfast or unpacking and putting away the contents of an office. Many of these tasks involve manipulating objects. We pile things up, put objects in boxes and drawers, and arrange them on shelves. Doing so requires an understanding of how the world works: depending on how the objects in a pile are arranged and what they are made of, a pile sometimes slips or falls over; pulling on a drawer usually opens it, but sometimes the drawer sticks; moving a box does not typically break the items inside it.

Building agents to perform these common tasks is a challenging problem. In this work, we approach the problem by developing a rule-based representation that such agents can use to model, and learn, the effects of acting on their environment. Learning allows the agents to adapt to new environments without requiring humans to hand-craft models, something humans are notoriously bad at, especially when numeric parametrization is required. The representation we use is both probabilistic and relational, and includes additional logical concepts. We present a supervised learning algorithm that uses this representation language to build a model of action effects given a set of example action executions. By optimizing the tradeoff between maximizing the likelihood of these examples and minimizing the complexity of the current hypothesis, this algorithm effectively selects a relational model structure, a set of model parameters, and a language of new relational concepts that together provide a compact, yet highly accurate description of action effects.

Any agent that hopes to act in the real world must be an integrated system that perceives the environment, understands, and commands motors to effect changes to it. Unfortunately, the current state of the art in reasoning, planning, learning, perception, locomotion, and manipulation is so far removed from human-level abilities, that we cannot yet contemplate working in an actual domain of interest. Instead, we choose to work in domains that are its almost ridiculously simplified proxies.[1]

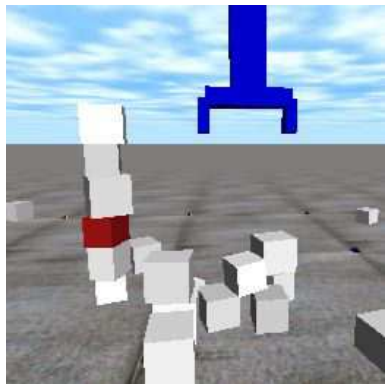

Figure 1: A three-dimensional blocks world simulation. The world consists of a table, several cubes of roughly uniform density but of varying size, and a robotic gripper that is moved by simulated motors.

One popular such proxy, used since the beginning of work in AI planning (Fikes & Nilsson, 1971) is a world of stacking blocks. It is typically formalized in some version of logic, using predicates such as $on(a, b)$ and $clear(a)$ to describe the relationships of the blocks to one another. Blocks are always very neatly stacked; they don't fall into jumbles. In this article, we present our work in the context of a slightly less ridiculous version of the blocks world, one constructed using a three-dimensional rigid-body dynamics simulator (ODE, 2004). An example world configuration is shown in Figure 1. In this simulated blocks world, blocks vary in size and colour; piles are not always tidy, and may sometimes fall over; and the gripper works only on medium-sized blocks, and is unreliable even there. Any approach capable of enabling effective behavior in this domain must handle its noisy, nondeterministic nature, and nontrivial dynamics, and so should then be able to handle other domains with similar characteristics.

One strategy for formulating such an approach is to learn models of the world's dynamics and then use them for planning different courses of action based on goals that may change over time. Another strategy is to assume a fixed goal or reward function, and to learn a policy that optimizes that reward function. In worlds of the complexity we are imagining, it would be impossible to establish, in advance, an appropriate reaction to every possible situation; in addition, we expect an agent to have an overall control architecture that is hierarchical, and for which an individual level in the hierarchy will have changing goals. For these reasons, we learn a model of the world dynamics, and then use it to make plans to achieve the goals at hand.

We begin this paper by describing the assumptions that underlie our modeling decisions. We then describe the syntax and semantics of our modeling language and give an algorithm for learning models in that language. To validate our models, we introduce a simple planning algorithm and then provide empirical results demonstrating the utility of the learned models by showing that we can plan with them. Finally, we survey relevant previous work, and draw our conclusions.

---

1. There is a very reasonable alternative approach, advocated by Brooks (1991), of working in the real world, with all its natural complexity, but solving problems that are almost ridiculously simplified proxies for the problems of interest.

## 2. Structured Stochastic Worlds

An agent introduced into a novel world must find the best possible explanation for the world's dynamics within the space of possible models it can represent, which is defined by the agent's representation language. The ideal language would be able to compactly model every action effect the agent might encounter, and no others. Any extra modeling capacity is wasted and will complicate learning, since the agent will have to consider a larger space of possible models, and be more likely to overfit its experience. Choosing a good representation language provides a strong *bias* for any algorithm that will learn models in that language.

Most languages that have been used to describe deterministic planning models are, at least on the surface, first order; that is, they abstract over the particular identities of objects, describing the effects of actions in terms of the properties of and relations among these objects. (They accomplish this by letting an action take arguments, and representing these arguments as variables.) This representational capacity is crucial for reasons of compactness and generalization: it is usually grossly inefficient to have to describe the behavior of individual objects.

Much of the original work on probabilistic planning uses the formalism of Markov decision processes, which represents the states of the world individually and atomically (Puterman, 1999). More recently, propositional (factored) representations of dynamics have been employed (Boyen & Koller, 1998; Guestrin, Koller, Parr, & Venkataraman, 2003), and some first-order representations have been developed, including probabilistic rules (Blum & Langford, 1999), equivalence classes (Draper, Hanks, & Weld, 1994), and the situation calculus approach of Boutilier, Reiter, and Price (2001). These representations also make it easy to articulate and take direct advantage of two useful assumptions about world dynamics: the *frame assumption*, which states that, when an agent takes an action in the world, anything not explicitly changed stays the same, and the *outcome assumption*, which states that each action affects the world in a small number of distinct ways, where each possible effect causes a set of changes to the world that happen together as a single *outcome*.

We take as our point of departure these probabilistic first-order representations of world dynamics. These representations have traditionally been applied to domains such as logistics planning or the traditional, abstract blocks world, which are idealized symbolic abstractions of an underlying domain. Our goal is to learn models of more realistic worlds, which requires us to adapt the modeling language to accommodate additional uncertainty and complexity. We do this by:

- Allowing rules to refer to objects not mentioned in the argument list of the action.

- Relaxing the frame assumption: allowing unmodeled "noise" changes in the world.

- Extending the language: allowing more complex forms of quantification and the construction of new concepts.

**Action parameterization** In traditional representations of action dynamics, any objects whose properties may be changed as the result of an action must be named in the argument list of the action. Instead, we define actions so that their parameters describe only those objects that are free parameters of the action: for example, the block to be picked up, and the object on which the currently held block is to be placed. However, actions can change the properties of other objects, ones that are not in their parameter list, and our models should have some way of determining which objects can be affected. In this paper, we introduce the use of *deictic references* to identify these objects. Deictic references (Agre & Chapman, 1987) identify objects relative to the agent or action being performed. For example, they can refer to objects such as the thing under the block to be picked up, the currently held object, or the table that the block accidentally falls onto. We use deictic references as a mechanism for adding new logical variables to our models, in much the same way as Benson (1996).

**Modeling noise** In complex domains, actions affect the world in a variety of ways. We must learn to model not only the circumstances under which they have reasonable effects, but also their behavior in unusual situations. This complicates the dynamics, and makes learning more difficult. Also, because these actions are executed in a physical world, they are not guaranteed to have a small number of simple effects, and as a result they may violate the outcomes assumption. In the blocks world this can happen, for example, when a stack is knocked over. We develop a simple noise mechanism that allows us to partially model action effects, ignoring ones that are rare or too complicated to model explicitly.

**Language extension** In traditional symbolic domains, rules are constructed using a predefined set of observable predicates. However, it is sometimes useful to define additional predicates whose truth values can be computed based on the predefined ones. This has been found to be essential for modeling certain advanced planning domains (Edelkamp & Hoffman, 2004).

In traditional blocks worlds, for example, the usual set of predicates contains *on*, *clear* and *inhand*. When working in our more realistic, noisy blocks world, we found that these predicates would not be sufficient to allow the agent to learn an accurate model. For example, it would be difficult to state that putting a block on a tall stack is likely to cause the stack to topple without having some concept of stack height, or to state that attempting to pick up a block that is not *clear* usually picks up the block on the top of its stack without having some way of describing the block on top of a stack.

While we could simply add those additional predicates that seem useful to the perceptual language, hand-engineering an appropriate language every time we tackle a new problem is difficult, time consuming, and error prone. State-of-the-art planning representations such as PDDL (Edelkamp & Hoffman, 2004) use a *concept language* to define new predicates or *concepts* in terms of previous, simpler ones. In this paper, we show that concepts can be learned, much like predicates can be invented in ILP (Khan, Muggleton, & Parson, 1998). As we will see, all of the traditional blocks world predicates, including *inhand* and *clear*, as well as other useful concepts such as *height*, are easily defined in terms of *on* given a simple concept language (Yoon, Fern, & Givan, 2002).

## 3. State and Action Representation

Our goal is to learn a model of the state transition dynamics of the world. To do so, we need to be able to represent the set $\mathcal{S}$ of possible states of the world and the set $\mathcal{A}$ of possible actions the agent can take. We represent both of these components using a subset of a relatively standard first-order logic with equality. The representation of states and actions is ground during inference, learning, and planning.

We begin by defining a primitive language which includes a set of constants $C$, a set of predicates $\Phi$, and a set of functions $\Omega$. There are three types of functions in $\Omega$: traditional functions, which range over objects; discrete-valued functions, which range over a predefined discrete set of values; and integer-valued functions, which range over a finite subset of the integers. All of these primitives can be observed directly in the world. (In this work, we assume that the environment is completely observable; that is, that the agent is able to perceive an unambiguous and correct description of the current state.[2]) The constants in $C$ can be assumed to have intrinsic meaning, or can be viewed as meaningless markers assigned by the perceptual system, as described in detail below.

### 3.1 State Representation

States describe all the possible different configurations of the properties of and relations between objects. Each state describes a particular configuration of these values for all of the objects in the world, where those individual objects are denoted using constants. There is no limit on the number of objects in a world configuration, though in our current formalism there is no mechanism for the creation or deletion of objects as a result of the world dynamics.

Formally, the state descriptions are conjunctive sentences of the form:

$$\bigwedge_{\phi \in \Phi} \bigwedge_{t \in G(C, m(\phi))} (\neg)\phi(t) \quad \wedge \quad \bigwedge_{\omega \in \Omega} \bigwedge_{t \in G(C, m(\omega))} \omega(t) = * \ ,$$

where $m(x)$ is the arity of predicate or function $x$, $C$ is the set $c_1 \ldots c_n$ of constants, $G(x, a)$ is the set of all length $a$ lists of elements from $x$, $(\neg)$ indicates that the predicates may be optionally negated, and $*$ indicates that functions can be assigned to any value in their range. In this manner, states list the truth values for all of the possible groundings of the predicates and functions with the terms. Such a sentence gives a complete specification, in the vocabulary of $\Phi$ and $\Omega$, of the properties and interrelations of the $|C|$ objects present in the world. (Note that predicate and function arguments are always constants, and never terms made using function symbols, so these descriptions are always finite given a finite language.)

In the rest of this section, we describe the two approaches to denoting objects using the constants in $C$, and illustrate them with example conjunctive state sentences.

#### 3.1.1 Intrinsic Constants

The first approach to state descriptions refers to objects using *intrinsic constants*. Each intrinsic constant is associated with a particular object and consistently used to denote that same object. Such constants are useful when the perceptual system has some unique way to identify perceived objects independent of their attributes and relations to one another. For example, an internet software agent might have access to universal identifiers that distinguish the objects it perceives.

---

2. This is a very strong, and ultimately indefensible assumption; one of our highest priorities for future work is to extend this to the case when the environment is partially observable.

As an example, let us consider representing the states of a simple blocks world, using a language that contains the predicates *on*, *clear*, *inhand*, *inhand-nil*, *block*, and *table*, and the integer-valued function *height*. The objects in this world include blocks BLOCK-A and BLOCK-B, a table TABLE, and a gripper. Blocks can be on other blocks or on the table. A block that has nothing on it is clear. The gripper can hold one block or be empty. The sentence

$$
\begin{aligned}
&inhand\text{-}nil \wedge on(\text{BLOCK-A, BLOCK-B}) \wedge on(\text{BLOCK-B, TABLE}) \wedge \neg on(\text{BLOCK-B, BLOCK-A}) \wedge \\
&\neg on(\text{BLOCK-A, TABLE}) \wedge \neg on(\text{TABLE, BLOCK-A}) \wedge \neg on(\text{TABLE, BLOCK-B}) \wedge \neg on(\text{TABLE, TABLE}) \wedge \\
&\neg on(\text{BLOCK-A, BLOCK-A}) \wedge \neg on(\text{BLOCK-B, BLOCK-B}) \wedge table(\text{TABLE}) \wedge \neg table(\text{BLOCK-A}) \wedge \\
&\neg table(\text{BLOCK-B}) \wedge block(\text{BLOCK-A}) \wedge block(\text{BLOCK-B}) \wedge \neg block(\text{TABLE}) \wedge clear(\text{BLOCK-A}) \wedge \\
&\neg clear(\text{BLOCK-B}) \wedge \neg clear(\text{TABLE}) \wedge \neg inhand(\text{BLOCK-A}) \wedge \neg inhand(\text{BLOCK-B}) \wedge \\
&\neg inhand(\text{TABLE}) \wedge height(\text{BLOCK-A}) = 2 \wedge height(\text{BLOCK-B}) = 1 \wedge height(\text{TABLE}) = 0
\end{aligned}
\tag{1}
$$

represents a blocks world where the gripper holds nothing and the two blocks are in a single stack on the table. Block BLOCK-A is on top of the stack, while BLOCK-B is below BLOCK-A and on the table TABLE.

Under this encoding, the sentence contains meaningful information about the objects' identities, which can then be used when learning about world dynamics.

### 3.1.2 SKOLEM CONSTANTS

Alternatively, states can also denote objects using *skolem constants*. Skolem constants are arbitrary identifiers that are associated with objects in the world but have no inherent meaning beyond how they are used in a state description.[3] Such constants are useful when the perceptual system has no way of assigning meaningful identifiers to the objects it observes. As an example, consider how a robot might build a state description of the room it finds itself in. We assume that this robot can observe the objects that are present, their properties, and their relationships to each other. However, when naming the objects, it has no reason to choose any particular name for any specific object. Instead, it just creates arbitrary identifiers, skolem constants, and uses them to build the state.

Using skolem constants, we can rewrite Sentence 1 as:

$$
\begin{aligned}
&inhand\text{-}nil \wedge on(c001, c002) \wedge on(c002, c004) \wedge \neg on(c002, c001) \wedge \\
&\neg on(c001, c004) \wedge \neg on(c004, c001) \wedge \neg on(c004, c002) \wedge \neg on(c004, c004) \wedge \\
&\neg on(c001, c001) \wedge \neg on(c002, c002) \wedge table(c004) \wedge \neg table(c001) \wedge \\
&\neg table(c002) \wedge block(c001) \wedge block(c002) \wedge \neg block(c004) \wedge clear(c001) \wedge \\
&\neg clear(c002) \wedge \neg clear(c004) \wedge \neg inhand(c001) \wedge \neg inhand(c002) \wedge \\
&\neg inhand(c004) \wedge height(c001) = 2 \wedge height(c002) = 1 \wedge height(c004) = 0
\end{aligned}
$$

Here, the perceptual system describes the table and the two blocks using the arbitrary constants *c004*, *c001*, and *c002*.

---

3. Skolem constants can be interpreted as skolemizations of existential variables.

From this perspective, states of the world are not isomorphic to interpretations of the logical language, since there might be many interpretations that satisfy a particular state-specification sentence; these interpretations will be the same up to permutation of the objects the constants refer to. This occurs because objects are only distinguishable based on their properties and relations to other objects.

The techniques we develop in this paper are generally applicable to representing and learning the dynamics of worlds with intrinsic constants or skolem constants. We will highlight the few cases where this is not true as they are presented. We will also see that the use of skolem constants is not only more perceptually plausible but also forces us to create new learning algorithms that abstract object identity more aggressively than previous work and can improve the quality of learned models.

### 3.2 Action Representation

Actions are represented as positive literals whose predicates are drawn from a special set, $\alpha$, and whose terms are drawn from the set of constants C associated with the world $s$ where the action is to be executed.

For example, in the simulated blocks world, $\alpha$ contains $pickup/1$, an action for picking up blocks, and $puton/1$, an action for putting down blocks. The action literal $pickup(\text{BLOCK-A})$ could represent the action where the gripper attempts to pickup the block BLOCK-A in the state represented in Sentence 1.

## 4. World Dynamics Representation

We are learning the probabilistic transition dynamics of the world, which can be viewed as the conditional probability distribution $\Pr(s'|s, a)$, where $s, s' \in \mathcal{S}$ and $a \in \mathcal{A}$. We represent these dynamics with rules constructed from the basic logic described in Section 3, using logical variables to abstract the identities of particular objects in the world.

In this section, we begin by describing a traditional representation of deterministic world dynamics. Next, we present the probabilistic case. Finally, we extend it in the ways we mentioned in Section 2: by permitting the rules to refer to objects not mentioned in the action description, by adding in noise, and by extending the language to allow for the construction of new concepts.

A *dynamic rule* for action $z$ has the form

$$\forall \bar{x}.\Psi(\bar{x}) \wedge z(\bar{x}) \rightarrow \bullet \Psi'(\bar{x}) \ ,$$

meaning that, for any vector of terms $\bar{x}$ such that the context $\Psi$ holds of them at the current time step, taking action $z(\bar{x})$ will cause the formula $\Psi'$ to hold of those same terms in the next step. The action $z(\bar{x})$ must contain every $x_i \in \bar{x}$. We constrain $\Psi$ and $\Psi'$ to be conjunctions of literals constructed from primitive predicates and terms $x_i \in \bar{x}$, or from functions applied to these terms and set equal to a value in their range. In addition, $\Psi$ is allowed to contain literals constructed from integer-valued functions of a term related to an integer in their range by greater-than or less-than predicates.

We will say that a rule *covers* a state $s$ and action $a$ if there exists a substitution $\sigma$ mapping the variables in $\bar{x}$ to C (note that there may be fewer variables in $\bar{x}$ than constants in C) such that $s \models \Psi(\sigma(\bar{x}))$ and $a = z(\sigma(\bar{x}))$. That is, there is a substitution of constants for variables that, when it is applied to the context $\Psi(\bar{x})$, grounds it so that it is entailed by the state $s$ and, when applied to the rule action $z(\bar{x})$, makes it equal to the action $a$.

Now, given that the rule covers $s$ and $a$, what can we say of the subsequent state $s'$? First, the rule directly specifies that $\Psi'(\sigma(\bar{x}))$ holds at the next step. But this may be only an incomplete specification of the state; we will use the frame assumption to fill in the rest:

$$s' = \Psi'(\sigma(\bar{x})) \quad \wedge \bigwedge_{\phi \in \Phi} \quad \bigwedge_{\{\{\phi(t):t\in G(C,m(\phi))\}-pos(\Psi'(\sigma(\bar{x})))\}} l(s, \phi, \sigma(\bar{x}))$$

$$\wedge \bigwedge_{\omega \in \Omega} \quad \bigwedge_{\{\{\omega(t):t\in G(C,m(\phi))\}-funct(\Psi'(\sigma(\bar{x})))\}} l(s, \omega, \sigma(\bar{x})) \quad,$$

where $l(s, y, t)$ stands for the literal in $s$ that has predicate or function $y$ and argument list $t$, $pos(\Psi')$ is the set of literals in $\Psi'$ with negations ignored, and $funct(\Psi')$ is the set of ground functions in $\Psi'$ extracted from their equality assignments. This is all to say that every literal that would be needed to make a complete description of the state but is not included in $\Psi'(\sigma(\bar{x}))$ is retrieved, with its associated truth value or equality assignment, from $s$.

In general we will have a set of rules for each action, but we will require their contexts to be mutually exclusive, so that any given state-action pair is covered by at most one rule; if it is covered by none, then we will assume that nothing changes. [4] As an example, consider a small set of rules for picking up blocks,

$$pickup(X, Y) : inhand\text{-}nil, on(X, Y), block(Y), height(Y) < 10$$
$$\rightarrow inhand(X), \neg on(X, Y), clear(Y),$$
$$pickup(X, Y) : inhand\text{-}nil, on(X, Y), table(Y)$$
$$\rightarrow inhand(X), \neg on(X, Y).$$

The top line of each rule shows the action followed by context; the next line describes the effects, or the outcome. According to these two rules, executing $pickup(X, Y)$ changes the world only when the hand is empty and when $X$ is on $Y$. The exact set of changes depends on whether $Y$ is the table, or a block of height nine or less.

## 4.1 Probabilistic Rules

The deterministic dynamics rules described above allow generalization over objects and exploitation of the frame assumption, but they are not very well suited for use in highly stochastic domains. In order to apply them to such domains we will have to extend them to describe the probability distribution over resulting states, $\Pr(s'|s, a)$. Probabilistic STRIPS operators (Blum & Langford, 1999) model how an agent's actions affect the world around it by describing how these actions alter the properties of and the relationships between objects in the world. Each rule specifies a small number of simple action *outcomes*—sets of changes that occur in tandem.

---

4. Without this restriction, we would need to define some method of choosing between the possibly conflicting predictions of the different covering rules. The simplest way to do so would involve picking one of the rules, perhaps the most specific one, or the one we are most confident of. (Rule confidence scores would have to be estimated.)

We can see such probabilistic rules as having the form

$$\forall \bar{x}.\Psi(\bar{x}) \wedge z(\bar{x}) \rightarrow \bullet \left\{ \begin{array}{ll} p_1 & \Psi'_1(\bar{x}) \\ \dots & \dots \\ p_n & \Psi'_n(\bar{x}) \end{array} \right. ,$$

where $p_1 \dots p_n$ are positive numbers summing to 1, representing a probability distribution, and $\Psi'_1 \dots \Psi'_n$ are formulas describing the subsequent state, $s'$.

Given a state $s$ and action $a$, we can compute coverage as we did in the deterministic case. Now, however, given a covering substitution $\sigma(\bar{x})$, probabilistic rules no longer predict a unique successor state. Instead, each $\Psi'_1 \dots \Psi'_n$ can be used to construct a new state, just as we did with the single $\Psi'$ in the deterministic case. There are $n$ such possible subsequent states, $s'_i$, each of which will occur with associated probability $p_i$.

The probability that a rule $r$ assigns to moving from state $s$ to state $s'$ when action $a$ is taken, $\Pr(s'|s,a,r)$, can be calculated as:

$$\begin{aligned} P(s'|s,a,r) &= \sum_{\Psi'_i \in r} P(s', \Psi'_i | s, a, r) \\ &= \sum_{\Psi'_i \in r} P(s'|\Psi'_i, s, a, r) P(\Psi'_i | s, a, r) \end{aligned} \quad (2)$$

where $P(\Psi'_i|s,a,r)$ is $p_i$, and the outcome distribution $P(s'|\Psi'_i, s, a, r)$ is a deterministic distribution that assigns all of its mass to the relevant $s'$. If $P(s'|\Psi'_i, s, a, r) = 1.0$, that is, if $s'$ is the state that would be constructed given that rule and outcome, we say that the outcome $\Psi'_i$ *covers* $s'$.

In general, it is possible, in this representation, for a subsequent state $s'$ to be covered by more than one of the rule's outcomes. In that case, the probability of $s'$ occurring is the sum of the probabilities of the relevant outcomes. Consider a rule for painting blocks:

$$paint(X) : inhand(X), block(X)$$
$$\rightarrow \left\{ \begin{array}{l} .8 : painted(X), wet \\ .2 : \text{no change.} \end{array} \right.$$

When this rule is used to model the transition caused by the action $paint(a)$ in an initial state that contains $wet$ and $painted(a)$, there is only one possible successor state: the one where no change occurs, so that $wet$ and $painted(a)$ remain true. Both the outcomes describe this one successor state, and so we must sum their probabilities to recover that state's total probability.

A set of rules specifies a complete conditional probability distribution $\Pr(s'|s,a)$ in the following way: if the current state $s$ and action $a$ are covered by exactly one rule, then the distribution over subsequent states is that prescribed by the rule. If not, then $s'$ is predicted to be the same as $s$ with probability 1.0.

As an example, a probabilistic set of rules for picking up blocks might look as follows:

$$pickup(X, Y) : inhand\text{-}nil, on(X, Y), block(Y), height(Y) < 10$$
$$\rightarrow \begin{cases} .7 : inhand(X), \neg on(X, Y), clear(Y) \\ .3 : \text{no change} \end{cases}$$
$$pickup(X, Y) : inhand\text{-}nil, on(X, Y), table(Y)$$
$$\rightarrow \begin{cases} .8 : inhand(X), \neg on(X, Y) \\ .2 : \text{no change} \end{cases}$$

The top line of each rule still shows the action followed by the context; the bracket surrounds the outcomes and their distribution. The outcomes are the same as before, only now there is a small chance that they will not occur.

## 4.2 Deictic Reference

In standard relational representations of action dynamics, a variable denoting an object whose properties may be changed as the result of an action must be named in the argument list of the action. This can result in awkwardness even in deterministic situations. For example, the abstract action of picking up a block must take two arguments. In $pickup(X, Y)$, $X$ is the block to be picked up and $Y$ is the block from which it is to be picked up. This relationship is encoded by an added condition $on(X, Y)$ in the rule's context. That condition does not restrict the applicability of the rule; it exists to guarantee that $Y$ is bound to the appropriate object. This restriction has been adopted because it means that, given a grounding of the action, all the variables in the rule are bound, and it is not necessary to search over all substitutions $\sigma$ that would allow a rule to cover a state. However, it can complicate planning because, in many cases, all ground instances of an operator are considered, even though most of them are eventually rejected due to violations of the preconditions. In our example, we would reject all instances violating the $on(X, Y)$ relation in the context.

In more complex domains, this requirement is even more awkward: depending on the circumstances, taking an action may affect different, varied sets of objects. In blocks worlds where a block may be *on* several others, the *pickup* action may affect the properties of each of those blocks. To model this without an additional mechanism for referring to objects, we might have to increase, or even vary, the number of arguments *pickup* takes.

To handle this more gracefully, we extend the rule formalism to include *deictic* references to objects. Each rule may be augmented with a list, $D$, of deictic references. Each deictic reference consists of a variable $V_i$ and restriction $\rho_i$ which is a set of literals that define $V_i$ with respect to the variables $\bar{x}$ in the action and the other $V_j$ such that $j < i$. These restrictions are supposed to pick out a single, unique object: if they do not—if they pick out several, or none—the rule fails to apply. So, to handle the *pickup* action described above, the action would have a single argument, $pickup(X)$, and the rule would contain a deictic variable $V$ with the constraint that $on(X, V)$.

To use rules with deictic references, we must extend our procedure for computing rule coverage to ensure that all of the deictic references can be resolved. The deictic variables may be bound simply by starting with bindings for $\bar{x}$ and working sequentially through the deictic variables, using their restrictions to determine their unique bindings. If at any point the binding for a deictic variable is not unique, it fails to refer, and the rule fails to cover the state—action pair.

This formulation means that extra variables need not be included in the action specification, which reduces the number of operator instances, and yet, because of the requirement for unique designation, a substitution can still be quickly discovered while testing coverage.

So, for example, to denote "the red block on the table" as $V_2$ (assuming that there were only one table and one such block) we would use the following deictic references:

$$V_1 \quad : \quad table(V_1)$$
$$V_2 \quad : \quad color(V_2) = red \wedge block(V_2) \wedge on(V_2, V_1) \ .$$

If there were several, or no, tables in the world, then, under our rule semantics, the first reference would fail: similarly, the second reference would fail if the number of red blocks on the unique table represented by $V_1$ were not one.

To give a more action-oriented example, when denoting "the block on top of the block I touched", where $touch(Z)$ was the action, we would use the following deictic reference:

$$V_1 \quad : \quad on(V_1, Z) \wedge block(V_1) \ .$$

A set of deictic probabilistic rules for picking up blocks might look as follows:

$$pickup(X) : \left\{ \ Y : inhand(Y), Z : table(Z) \ \right\}$$
empty context
$$\rightarrow \left\{ \begin{array}{l} .9 : inhand\text{-}nil, \neg inhand(Y), on(Y, Z) \\ .1 : \text{no change} \end{array} \right.$$

$$pickup(X) : \left\{ \ Y : block(Y), on(X, Y) \ \right\}$$
$$inhand\text{-}nil, height(Y) < 10$$
$$\rightarrow \left\{ \begin{array}{l} .7 : inhand(X), \neg on(X, Y), clear(Y) \\ .3 : \text{no change} \end{array} \right.$$

$$pickup(X) : \left\{ \ Y : table(Y), on(X, Y) \ \right\}$$
$$inhand\text{-}nil$$
$$\rightarrow \left\{ \begin{array}{l} .8 : inhand(X), \neg on(X, Y) \\ .2 : \text{no change} \end{array} \right.$$

The top line of each rule now shows the action followed by the deictic variables, where each variable is annotated with its restriction. The next line is the context, and the outcomes and their distribution follow. The first rule applies in situations where there is something in the gripper, and states that there is a probability of 0.9 that action will cause the gripped object to fall to the table, and that nothing will change otherwise. The second rule applies in situations where the object to be picked up is on another block, and states that the probability of success is 0.7. The third rule applies in situations where the object to be picked up is on the table and describes a slightly higher success probability, 0.8. Note that different objects are affected, depending on the state of the world.

### 4.3 Adding Noise

Probability models of the type we have seen thus far, ones with a small set of possible outcomes, are not sufficiently flexible to handle the noise of the real world. There may be a large number of possible outcomes that are highly unlikely, and reasonably hard to model: for example, all the configurations that may result when a tall stack of blocks topples. It would be inappropriate to model such outcomes as impossible, but we don't have the space or inclination to model each of them as an individual outcome.

So, we will allow our rule representation to account for some results as noise. By definition, noise will be able to represent the outcomes whose probability we haven't quantified. Thus, by allowing noise, we will lose the precision of having a true probability distribution over next states.

To handle noise, we must change our rules in two ways. First, each rule will have an additional *noise outcome* $\Psi'_{noise}$, with an associated probability $P(\Psi'_{noise}|s, a, r)$; now, the set of outcome probabilities that must sum to 1.0 will include $P(\Psi'_{noise}|s, a, r)$ as well as $P(\Psi'_1|s, a, r) \ldots P(\Psi'_n|s, a, r)$. However, $\Psi'_{noise}$ will not have an associated list of literals, since we are declining to model in detail what happens to the world in such cases.

Second, we will create an additional *default rule*, with an empty context and two outcomes: an empty outcome (which, in combination with the frame assumption, models the situations where nothing changes), and, again, a noise outcome (modeling all other situations). This rule allows noise to occur in situations where no other specific rule applies; the probability assigned to the noise outcome in the default rule specifies a kind of "background noise" level.

Since we are not explicitly modeling the effects of noise, we can no longer calculate the transition probability $\Pr(s'|s, a, r)$ using Equation 2: we lack the required distribution $P(s'|\Psi'_i, s, a, r)$ for the noise outcome. Instead, we substitute a worst case constant bound $p_{min} \leq P(s'|\Psi'_{noise}, s, a, r)$. This allows us to bound the transition probability as

$$
\begin{aligned}
\hat{P}(s'|s, a, r) &= p_{min}P(\Psi'_{noise}|s, a, r) + \sum_{\Psi'_i \in r} P(s'|\Psi'_i, s, a, r)P(\Psi'_i|s, a, r) \\
&\leq P(s'|s, a, r).
\end{aligned}
\tag{3}
$$

Intuitively, $p_{min}$ assigns a small amount of probability mass to every possible next state $s'$. Note that it can take a value higher than the true minimum: it is an approximation. However, to ensure that the probability model remains well-defined, $p_{min}$ times the number of possible states should not exceed 1.0.

In this way, we create a partial model that allows us to ignore unlikely or overly complex state transitions while still learning and acting effectively. [5]

Since these rules include noise and deictic references, we call them *Noisy Deictic Rules* (NDRs). In a rather stochastic world, the set of NDRs for picking up blocks might now

---

5. $P(s'|\Psi'_{noise}, s, a, r)$ could be modeled using any well-defined probability distribution describing the noise of the world, which would give us a full distribution over the next states. The premise here is that it might be difficult to specify such a distribution–in our domain, we would have to ensure that this distribution does not assign probability to worlds that are impossible, such as worlds where some blocks are floating in midair. As long as these events are unlikely enough that we would not want to consider them while planning, it is reasonable to not model them directly.

look as follows:

$$pickup(X) : \left\{ \; Y : inhand(Y), Z : table(Z) \; \right\}$$

empty context

$$\rightarrow \begin{cases} .6 : inhand\text{-}nil, \neg inhand(Y), on(Y, Z) \\ .1 : \text{no change} \\ .3 : \text{noise} \end{cases}$$

$$pickup(X) : \left\{ \; Y : block(Y), on(X, Y) \; \right\}$$

$inhand\text{-}nil, height(Y) < 10$

$$\rightarrow \begin{cases} .7 : inhand(X), \neg on(X, Y), clear(Y) \\ .1 : \text{no change} \\ .2 : \text{noise} \end{cases}$$

$$pickup(X) : \left\{ \; Y : table(Y), on(X, Y) \; \right\}$$

$inhand\text{-}nil$

$$\rightarrow \begin{cases} .8 : inhand(X), \neg on(X, Y) \\ .1 : \text{no change} \\ .1 : \text{noise} \end{cases}$$

default rule:

$$\rightarrow \begin{cases} .6 : \text{no change} \\ .4 : \text{noise} \end{cases}$$

The format of the rules is the same as before, in Section 4.2, except that each rule now includes an explicit noise outcome. The first three rules are very similar to their old versions, the only difference being that they model noise. The final rule is the default rule: it states that, if no other rule applies, the probability of observing a change is 0.4.

Together these rules provide a complete example of the type of rule set that we will learn in Section 5.1. However, they were written with a fixed modeling language of functions and predicates. The next section describes how concepts can be used to extend this language.

## 4.4 Concept Definitions

In addition to the observed primitive predicates, it is often useful to have background knowledge that defines additional predicates whose truth values can be computed based on the observations. This has been found to be essential for modeling certain planning domains (Edelkamp & Hoffman, 2004).

This background knowledge consists of definitions for additional *concept predicates and functions*. In this work, we express concept definitions using a *concept language* that includes conjunction, existential quantification, universal quantification, transitive closure, and counting. Quantification is used for defining concepts such as $inhand(X) := block(X) \wedge \neg \exists Y. on(X, Y)$. Transitive closure is included in the language via the Kleene star operator and defines concepts such as $above(X, Y) := on^*(X, Y)$. Finally, counting is included using a special quantifier $\#$ which returns the number of objects for which a formula is true. It is useful for defining integer-valued functions such as $height(X) := \#Y. above(X, Y)$.

Once defined, concepts enable us to simplify the context and the deictic variable definitions, as well as to restrict them in ways that cannot be described using simple conjunctions. Note, however, that there is no need to track concept values in the outcomes, since they can always be computed from the primitives. Therefore, only the rule contexts use a language enriched by concepts; the outcomes contain primitives.

As an example, here is a deictic noisy rule for attempting to pick up block $X$, side by side with the background knowledge necessary when the only primitive predicates are *on* and *table*.

$$
pickup(X) : \left\{ \begin{array}{l} Y : topstack(Y, X), \\ Z : on(Y, Z), \\ T : table(T) \end{array} \right\}
$$

$$
inhand\text{-}nil, height(Y) < 9
$$

$$
\rightarrow \left\{ \begin{array}{ll} .80 : & \neg on(Y, Z) \\ .10 : & \neg on(Y, Z), on(Y, T) \\ .05 : & \text{no change} \\ .05 : & \text{noise} \end{array} \right.
$$

$$
\begin{array}{rcl}
clear(X) & := & \neg \exists Y.on(Y, X) \\
inhand(X) & := & block(X) \wedge \neg \exists Y.on(X, Y) \\
inhand\text{-}nil & := & \neg \exists Y.inhand(Y) \\
above(X, Y) & := & on^*(X, Y) \\
topstack(X, Y) & := & clear(X) \wedge above(X, Y) \\
height(X) & := & \# Y.above(X, Y)
\end{array}
\tag{4}
$$

The rule is more complicated than the example rules given thus far: it deals with the situation where the block to be picked up, $X$, is in the middle of a stack. The deictic variable $Y$ identifies the (unique) block on top of the stack, the deictic variable $Z$—the object under $Y$, and the deictic variable $T$—the table. As might be expected, the gripper succeeds in lifting $Y$ with a high probability.

The concept definitions include $clear(X)$, defined as "There exists no object that is on $X$"; $inhand(X)$, defined as "$X$ is a block that is not on any object"; $inhand\text{-}nil$, defined as "There exists no object such that it is in the hand"; $above(X, Y)$, defined as the transitive closure of $on(X, Y)$; $topstack(X, Y)$, defined as "$X$ is above $Y$, and clear"; and $height(X)$, defined as "The number of objects that can are below $X$ using a chain of $on$s." As explained above, these concepts are used only in the context and the deictic variable definitions, while outcomes track only the primitive predicates; in fact, only $on$ appears in the outcomes, since the value of the *table* predicates never changes.

## 4.5 Action Models

We combine a set of concept definitions and a set of rules to define an *action model*. Our best action models will represent the rule set using NDRs, but, for comparison purposes, some of our experiments will involve rule sets that use simpler representations, without noise or deictic references. Moreover, the rule sets will differ in whether they are allowed to contain constants. The rules presented so far have contained none, neither in their context nor in the outcomes. This is the only reasonable setup when states contain skolem constants, as these constants have no inherent meaning and the names they are assigned will not in general be repeated. However, when states have intrinsic constants, it is perfectly acceptable to include constants in action models. After all, these constants can be used to uniquely identify objects in the world.

As we develop a learning algorithm in the next section, we will assume in general that constants are allowed in the action model, but we will show how simple restrictions within the algorithm can ensure that the learned models do not contain any. We also show, in Section 7, that learning action models which are restricted to be free of constants provides a useful bias that can improve generalization when training with small data sets.

## 5. Learning Action Models

Now that we have defined rule action models, we will describe how they may be constructed using a learning algorithm that attempts to return the action model that best explains a set of example actions and their results. More formally, this algorithm takes a training set $E$, where each example is a $(s, a, s')$ triple, and searches for an action model $A$ that maximizes the likelihood of the action effects seen in $E$, subject to a penalty on complexity.

Finding $A$ involves two distinct problems: defining a set of concept predicates, and constructing a rule set $R$ using a language that contains these predicates together with the directly observable primitive predicates. In this section, we first discuss the second problem, rule set learning, assuming some fixed set of predicates is provided to the learner. Then, we present a simple algorithm that discovers new, useful concept predicates.

### 5.1 Learning Rule Sets

The problem of learning rule sets is, in general, NP-hard (Zettlemoyer, Pasula, & Kaelbling, 2003). Here, we address this problem by using greedy search. We structure the search hierarchically by identifying two self-contained subproblems: outcome learning, which is a subproblem of the general rule set search, and parameter estimation, which is a subproblem of outcome learning. Thus, the overall algorithm involves three levels of greedy search: an outermost level, *LearnRules*, which searches through the space of rule sets, often by constructing new rules, or altering existing ones; a middle level, *InduceOutcomes* which, given an incomplete rule consisting of a context, an action, and a set of deictic references, fills in the rest of the rule; and an innermost level, *LearnParameters*, which takes a slightly more complete rule, now lacking only a distribution over the outcomes, and finds the distribution that optimizes the likelihood of the examples covered by this rule. We present these three levels starting from the inside out, so that each subroutine is described before the one that depends on it. Since all three subroutines attempt to maximize the same scoring metric, we begin by introducing this metric.

#### 5.1.1 The Scoring Metric

A greedy search algorithm must judge which parts of the search space are the most desirable. Here, this is done with the help of a scoring metric over rule sets,

$$S(R) = \sum_{(s,a,s') \in E} \log(\hat{P}(s'|s, a, r_{(s,a)})) - \alpha \sum_{r \in R} PEN(r) \tag{5}$$

where $r_{(s,a)}$ is the rule governing the transition occurring when $a$ is performed in $s$, $\alpha$ is a scaling parameter, and $PEN(r)$ is a complexity penalty applied to rule $r$. Thus, $S(R)$ favors rule sets that maximize the likelihood bound on the data and penalizes rule sets that are overly complex.

Ideally, $\hat{P}$ would be the likelihood of the example. However, rules with noise outcomes cannot assign an exact likelihood so, in their case, we use the lower bound defined in Equa-tion 3 instead. $PEN(r)$ is defined simply as the total number of literals in $r$. We chose this penalty for its simplicity, and also because it performed no worse than any other penalty term we tested in informal experiments. The scaling parameter $\alpha$ is set to 0.5 in our experiments, but it could also be set using cross-validation on a hold-out dataset or some other principled technique. This metric puts pressure on the model to explain examples using non-noise outcomes, which increases $\hat{P}$, but also has opposing pressure on complexity, via $PEN(r)$.

If we assume that each state-action pair $(s, a)$ is covered by at most one rule (which, for any finite set of examples, can be enforced simply by ensuring that each example's state-action pair is covered by at most one rule) we can rewrite the metric in terms of rules rather than examples, to give

$$S(R) \;=\; \sum_{r \in R} \sum_{(s,a,s') \in E_r} \log(\hat{P}(s'|s,a,r)) - \alpha PEN(r) \qquad (6)$$

where $E_r$ is the set of examples covered by $r$. Thus, each rule's contribution to $S(R)$ can be calculated independently of the others'.

### 5.1.2 Learning Parameters

The first of the algorithms described in this section, *LearnParameters*, takes an incomplete rule $r$ consisting of an action, a set of deictic references, a context, and a set of outcomes, and learns the distribution $P$ that maximizes $r$'s score on the examples $E_r$ covered by it.

Since the procedure is not allowed to alter the number of literals in the rule, and therefore cannot affect the complexity penalty term, the optimal distribution is simply the one that maximizes the log likelihood of $E_r$. In the case of rules with noise outcomes this will be

$$
\begin{aligned}
L \;&=\; \sum_{(s,a,s') \in E_r} \log(\hat{P}(s'|s,a,r)) \\
&=\; \sum_{(s,a,s') \in E_r} \log \left( p_{min} P(\Psi'_{noise}|s,a,r) + \sum_{\Psi'_i \in r} P(s'|\Psi'_i, s, a, r) P(\Psi'_i|s,a,r) \right) . \quad (7)
\end{aligned}
$$

For each non-noise outcome, $P(s'|\Psi'_i, s, a, r)$ is one if $\Psi'_i$ covers $(s, a, s')$ and zero otherwise. (In the case of rules without noise outcomes, the sum will be slightly simpler, with the $p_{min} P(\Psi'_{noise}|s,a,r)$ term missing.)

When every example is covered by a unique outcome, $L$'s maximum can be expressed in a closed form. Let the set of examples covered by an outcome $\Psi'$ be $E_{\Psi'}$. If we add a Lagrange multiplier to enforce the constraint that the $P(\Psi'_i|s,a,r)$ distributions must sum to 1.0, we will get

$$
\begin{aligned}
L \;&=\; \sum_{(s,a,s') \in E_r} \log \left( \sum_{\Psi'_i \in r} P(s'|\Psi'_i, s, a, r) P(\Psi'_i|s,a,r) \right) + \lambda (\sum_{\Psi'_i} P(\Psi'_i|s,a,r) - 1.0) \\
&=\; \sum_{E_{\Psi'}} |E_{\Psi'}| \log \left( P(\Psi'_i|s,a,r) \right) + \lambda (\sum_{\Psi'_i} P(\Psi'_i|s,a,r) - 1.0) .
\end{aligned}
$$

Then, the partial derivative of $L$ with respect to $P(\Psi'_i|s, a, r)$ will be $|E_{\Psi'}|/P(\Psi'_i|s, a, r) - \lambda$ and $\lambda = |E|$, so that $P(\Psi'_i|s, a, r) = |E_{\Psi'}|/|E|$. Thus, the parameters can be estimated by calculating the percentage of the examples that each outcome covers.

However, as we have seen in Section 4.1, it is possible for each example to be covered by more than one outcome; indeed, when we have a noise outcome, which covers all examples, this will always be the case. In this situation, the sum over examples cannot be rewritten as a simple sum of terms each representing a different outcomes and containing only a single relevant probability: the probabilities of overlapping outcomes remain tied together, no general closed-form solution exists, and estimating the maximum-likelihood parameters is a nonlinear programming problem. Fortunately, it is an instance of the well-studied problem of maximizing a concave function (the log likelihood presented in Equation 7) over a probability simplex. Several gradient ascent algorithms are known for this problem (Bertsekas, 1999); since the function is concave, they are guaranteed to converge to the global maximum. *LearnParameters* uses the *conditional gradient method*, which works by, at each iteration, moving along the parameter axis with the maximal partial derivative. The step-sizes are chosen using the Armijo rule (with the parameters $s = 1.0$, $\beta = 0.1$, and $\sigma = 0.01$.) The search converges when the improvement in $L$ is very small, less than $10^{-6}$. We chose this algorithm because it was easy to implement and converged quickly for all of the experiments that we tried. However, if problems are found where this method converges too slowly, one of the many other nonlinear optimization methods, such as a constrained Newton's method, could be directly applied.

### 5.1.3 Inducing Outcomes

Given *LearnParameters*, an algorithm for learning a distribution over outcomes, we can now consider the problem of taking an incomplete rule $r$ consisting of a context, an action, and perhaps a set of deictic references, and finding the optimal way to fill in the rest of the rule—that is, the set of outcomes $\{\Psi'_1 \ldots \Psi'_n\}$ and the associated distribution $P$ that maximize the score

$$S(r) \;\; = \sum_{(s,a,s') \in E_r} \log(\hat{P}(s'|s, a, r)) - \alpha PEN_o(r),$$

where $E_r$ is the set of examples covered by $r$, and $PEN_o(r)$ is the total number of literals in the outcomes of $r$. ($S(r)$ is that factor of the scoring metric in Equation 6 which is due to rule $r$, without those aspects of $PEN(r)$ which are fixed for the purposes of this subroutine: the number of literals in the context.)

In general, outcome induction is NP-hard (Zettlemoyer, Pasula, & Kaelbling, 2003). *InduceOutcomes* uses greedy search through a restricted subset of possible outcome sets: those that are *proper* on the training examples, where an outcome set is proper if every outcome covers at least one training example. Two operators, described below, move through this space until there are no more immediate moves that improve the rule score. For each set of outcomes it considers, *InduceOutcomes* calls *LearnParameters* to supply the best $P$ it can.

The initial set of outcomes is created by, for each example, writing down the set of atoms that changed truth values as a result of the action, and then creating an outcome to describe every set of changes observed in this way.

$$E_1 = t(c1), h(c2) \rightarrow h(c1), h(c2) \qquad \Psi'_1 = \{h(c1)\}$$
$$E_2 = h(c1), t(c2) \rightarrow h(c1), h(c2) \qquad \Psi'_2 = \{h(c2)\}$$
$$E_3 = h(c1), h(c2) \rightarrow t(c1), t(c2) \qquad \Psi'_3 = \{t(c1), t(c2)\}$$
$$E_4 = h(c1), h(c2) \rightarrow h(c1), h(c2) \qquad \Psi'_4 = \{\text{no change}\}$$

(a) $\qquad\qquad\qquad\qquad\qquad\qquad$ (b)

Figure 2: (a) Possible training data for learning a set of outcomes. (b) The initial set of outcomes that would be created from the data in (a) by picking the "smallest" outcome that describes each change.

As an example, consider the coins domain. Each coins world contains $n$ coins, which can be showing either heads or tails. The action *flip-coupled*, which takes no arguments, flips all of the coins, half of the time to heads, and otherwise to tails. A set of training data for learning outcomes with two coins might look like part (a) of Figure 2 where $h(C)$ stands for $heads(C)$, $t(C)$ stands for $\neg heads(C)$, and $s \rightarrow s'$ is part of an $(s, a, s')$ example where $a = $ *flip-coupled*. Now suppose that we have suggested a rule for *flip-coupled* that has no context or deictic references. Given our data, the initial set of outcomes has the four entries in part (b) of Figure 2.

If our rule contained variables, either as abstract action arguments or in the deictic references, *InduceOutcomes* would introduce those variables into the appropriate places in the outcome set. This variable introduction is achieved by applying the inverse of the action substitution to each example's set of changes while computing the initial set of outcomes. [6] So, given a deictic reference $C : red(C)$ which was always found to refer to $c1$, the only red coin, our example set of outcomes would contain $C$ wherever it currently contains $c1$.

Finally, if we disallow the use of constants in our rules, variables become the only way for outcomes to refer to the objects whose properties have changed. Then, changes containing a constant which is not referred to by any variable cannot be expressed, and the corresponding example will have to be covered by the noise outcome.

OUTCOME SEARCH OPERATORS

*InduceOutcomes* uses two search operators. The first is an *add* operator, which picks a pair of non-contradictory outcomes in the set and creates a new outcome that is their conjunction. For example, it might pick $\Psi'_1$ and $\Psi'_2$ and combine them, adding a new outcome $\Psi'_5 = \{h(c1), h(c2)\}$ to the set. The second is a *remove* operator that drops an outcome from the set. Outcomes can only be dropped if they were overlapping with other outcomes on every example they cover, otherwise the outcome set would not remain proper. (Of course, if the outcome set contains a noise outcome, then every other outcome can be dropped, since all of its examples are covered by the noise outcome.) Whenever an operator adds or removes an outcome, *LearnParameters* is called to find the optimal distribution over the new outcome set, which can then be used to calculate the maximum log likelihood of the data with respect to the new outcome set.

---

6. Thus, *InduceOutcomes* introduces variables aggressively wherever possible, based on the intuition that if any of the corresponding objects would be better described by a constant, this will become apparent through some other training example.

Sometimes, *LearnParameters* will return zero probabilities for some of the outcomes. Such outcomes are removed from the outcome set, since they contribute nothing to the likelihood, and only add to the complexity. This optimization improves the efficiency of the search.

In the outcomes of Figure 2, $\Psi'_4$ can be dropped since it covers only $E_4$, which is also covered by both $\Psi'_1$ and $\Psi'_2$. The only new outcome that can be created by conjoining the existing ones is $\Psi'_5 = \{h(c1), h(c2)\}$, which covers $E_1$, $E_2$, and $E_3$. Thus, if $\Psi'_5$ is added, then $\Psi'_1$ and $\Psi'_2$ can be dropped. Adding $\Psi'_5$ and dropping $\Psi'_1$, $\Psi'_2$, and $\Psi'_4$ creates the outcome set $\{\Psi'_3, \Psi'_5\}$, which is the optimal set of outcomes for the training examples in Figure 2.

Notice that an outcome is always equal to the union of the sets of literals that change in the training examples it covers. This fact ensures that every proper outcome can be made by merging outcomes from the initial outcome set. *InduceOutcomes* can, in theory, find any set of outcomes.

### 5.1.4 Learning Rules

Now that we know how to fill in incomplete rules, we will describe *LearnRules*, the outermost level of our learning algorithm, which takes a set of examples **E** and a fixed language of primitive and derived predicates, and performs a greedy search through the space of rule sets. More precisely, it searches through the space of *proper* rule sets, where a rule set $R$ is defined as proper with respect to a data set $E$ if it includes at most one rule that is applicable to every example $e \in E$ in which some change occurs, and if it does not include any rules that are applicable to no examples.

The search proceeds as described in the pseudocode in Figure 3. It starts with a rule set that contains only the default rule. At every step, it takes the current rule set and applies all its search operators to it to obtain a set of new rule sets. It then selects the rule set $R$ that maximizes the scoring metric $S(R)$ as defined in Equation 5. Ties in $S(R)$ are broken randomly.

We will begin by explaining how the search is initialized, then go on to describe the operators used, and finish by working through a simple example that shows *LearnRules* in action.

#### Rule Set Search Initialization

*LearnRules* can be initialized with any proper rule set. In this paper, we always initialize the set with only the noisy default rule. This treats all action effects in the training set as noise; as the search progresses, the search operators will introduce rules to explain action effects explicitly. We chose this initial starting point for its simplicity, and because it worked well in informal experiments. Another strategy would be to start with a very specific rule set, describing in detail all the examples. Such bottom-up methods have the advantage of being data-driven, which can help search reach good parts of the search space more easily. However, as we will show, several of the search operators used by the algorithm presented here are guided by the training examples, so the algorithm already has this desirable property. Moreover, this bottom-up method has bad complexity properties in

**LearnRuleSet(E)**
**Inputs:**
   Training examples **E**
**Computation:**
   Initialize rule set $R$ to contain only the default rule
   While better rules sets are found
     For each search operator $O$
       Create new rule sets with $O$, $R_O = O(R, \mathbf{E})$
       For each rule set $R' \in R_O$
         If the score improves $(S(R') > S(R))$
           Update the new best rule set, $R = R'$
**Output:**
   The final rule set $R$

Figure 3: *LearnRuleSet* pseudocode. This algorithm performs greedy search through the space of rule sets. At each step a set of search operators each propose a set of new rule sets. The highest scoring rule set is selected and used in the next iteration.

situations where a large data set can be described using a relatively simple set of rules, which is the case we are most interested in.

RULE SET SEARCH OPERATORS

During rule set search, *LearnRules* repeatedly finds and applies the operator that will increase the score of the current rule set the most.

Most of the search operators work by creating a new rule or set of rules (usually by altering an existing rule) and then integrating these new rules into the rule set in a way that ensures the rule set remains proper. Rule creation involves picking an action $z$, a set of deictic references $D$, and a context $\Psi$, and then calling on the *InduceOutcomes* learning algorithm to complete the rule by filling in the $\Psi_i'$s and $p_i$s. (If the new rule covers no examples, the attempt is abandoned, since adding such a rule cannot help the scoring metric.) Integration into a rule set involves not just adding the new rules, but also removing the old rules that cover any of the same examples. This can increase the number of examples covered by the default rule.

### 5.1.5 SEARCH OPERATORS

Each search operator $O$ takes as input a rule set $R$ and a set of training examples **E**, and creates a set of new rule sets $R_O$ to be evaluated by the greedy search loop. There are eleven search operators. We first describe the most complex operator, *ExplainExamples*, followed by the most simple one, *DropRules*. Then, we present the remaining nine operators, which all share the common computational framework outlined in Figure 4.

Together, these operators provide many different ways of moving through the space of possible rule sets. The algorithm can be adapted to learn different types of rule sets (for example, with and without constants) by restricting the set of search operators used.

**OperatorTemplate**$(R, \mathbf{E})$
**Inputs:**
   Rule set $R$
   Training examples $\mathbf{E}$
**Computation:**
   Repeatedly select a rule $r \in R$
     Create a copy of the input rule set $R' = R$
     Create a new set of rules, $N$, by making changes to $r$
     For each new rule $r' \in N$ that covers some examples
       Estimate new outcomes for $r'$ with *InduceOutcomes*
       Add $r'$ to $R'$ and remove any rules in $R'$ that cover any
         examples $r'$ covers
     Recompute the set of examples that the default rule in $R'$
       covers and the parameters of this default rule
     Add $R'$ to the return rule sets $R_O$
**Output:**
   The set of rules sets, $R_O$

Figure 4: *OperatorTemplate* Pseudocode. This algorithm is the basic framework that is used by six different search operators. Each operator repeatedly selects a rule, uses it to make $n$ new rules, and integrates those rules into the original rule set to create a new rule set.

- *ExplainExamples* takes as input a training set $\mathbf{E}$ and a rule set $R$ and creates new, alternative rule sets that contain additional rules modeling the training examples that were covered by the default rule in $R$. Figure 5 shows the pseudocode for this algorithm, which considers each training example $E$ that was covered by the default rule in $R$, and executes a three-step procedure. The first step builds a large and specific rule $r'$ that describes this example; the second step attempts to trim this rule, and so generalize it so as to maximize its score, while still ensuring that it covers $E$; and the third step creates a new rule set $R'$ by copying $R$ and integrating the new rule $r'$ into this new rule set.

  As an illustration, let us consider how steps 1 and 2 of *ExplainExamples* might be applied to the training example $(s, a, s') = (\{on(a, t), on(b, a)\}, pickup(b), \{on(a, t)\})$, when the background knowledge is as defined for Rule 4 in Section 4.4 and constants are not allowed.

  Step 1 builds a rule $r$. It creates a new variable $X$ to represent the object $b$ in the action; then, the action substitution becomes $\sigma = \{X \rightarrow b\}$, and the action of $r$ is set to $pickup(X)$. The context of $r$ is set to the conjunction *inhand-nil*, $\neg inhand(X)$, $clear(X)$, $height(X) = 2$, $\neg on(X, X)$, $\neg above(X, X)$, $\neg topstack(X, X)$. Then, in Step 1.2, *ExplainExamples* attempts to create deictic references that name the constants whose properties changed in the example, but which are not already in the action substitution. In this case, the only changed literal is $on(b, a)$, and $b$ is in the substitution, so $C = \{a\}$; a new deictic variable $Y$ is created and restricted, and $\sigma$ is extended to

**ExplainExamples**($R$, **E**)
**Inputs:**
    A rule set $R$
    A training set **E**
**Computation:**
    For each example $(s, a, s') \in$ **E** covered by the default rule in $R$
        **Step 1:** *Create a new rule* $r$
            **Step 1.1:** *Create an action and context for* $r$
                Create new variables to represent the arguments of $a$
                Use them to create a new action substitution $\sigma$
                Set $r$'s action to be $\sigma^{-1}(a)$
                Set $r$'s context to be the conjunction of boolean and equality literals that can
                  be formed using the variables and the available functions and predicates
                  (primitive and derived) and that are entailed by $s$
            **Step 1.2:** *Create deictic references for* $r$
                Collect the set of constants $C$ whose properties changed from $s$ to $s'$, but
                  which are not in $\sigma$
                For each $c \in C$
                    Create a new variable $v$ and extend $\sigma$ to map $v$ to $c$
                    Create $\rho$, the conjunction of literals containing $v$ that can be formed using
                      the available variables, functions, and predicates, and that are entailed by $s$
                    Create deictic reference $d$ with variable $v$ and restriction $\sigma^{-1}(\rho)$
                    If $d$ uniquely refers to $c$ in $s$, add it to $r$
            **Step 1.3:** *Complete the rule*
                Call *InduceOutcomes* to create the rule's outcomes.
        **Step 2:** *Trim literals from* $r$
            Create a rule set $R'$ containing $r$ and the default rule
            Greedily trim literals from $r$, ensuring that $r$ still covers $(s, a, s')$ and filling in the
              outcomes using *InduceOutcomes* until $R'$'s score stops improving
        **Step 3:** *Create a new rule set containing* $r$
            Create a new rule set $R' = R$
            Add $r$ to $R'$ and remove any rules in $R'$ that cover any examples $r$ covers
            Recompute the set of examples that the default rule in $R'$ covers and the parameters
              of this default rule
            Add $R'$ to the return rule sets $R_O$
**Output:**
    A set of rule sets, $R_O$

Figure 5: *ExplainExamples* Pseudocode. This algorithm attempts to augment the rule set with new
           rules covering examples currently handled by the default rule.

be $\{X \to b, Y \to a\}$. Finally, in Step 1.3, the outcome set is created. Assuming that of the examples for which context applies, nine out of ten end with $X$ being lifted, and the rest with it falling onto the table, the resulting rule $r'$ looks as follows:

$$pickup(X) : \left\{ Y : \begin{array}{l} \neg inhand(Y), \neg clear(Y), on(X, Y), \neg table(Y) \\ above(X, Y), topstack(X, Y), \neg above(Y, Y) \\ \neg topstack(Y, Y), \neg on(Y, Y), height(Y) = 1 \end{array} \right\}$$

$$inhand\text{-}nil, \neg inhand(X), clear(X), \neg table(X), height(X) = 2, \neg on(X, X),$$
$$\neg above(X, X), \neg topstack(X, X)$$
$$\to \left\{ \begin{array}{l} 0.9 : \neg on(X, Y) \\ 0.1 : \text{noise} \end{array} \right.$$

(The "falls on table" outcome is modeled as noise, since in the absence of constants the rule has no way of referring to the table.)

In Step 2, *ExplainExamples* trims this rule to remove the literals that are always true in the training examples, like $\neg on(X, X)$, and the $\neg table()$s, and the redundant ones, like $\neg inhand()$, $\neg clear(Y)$, and perhaps one of the *height*s, to give

$$pickup(X) : \left\{ Y : on(X, Y) \right\}$$
$$inhand\text{-}nil, clear(X), height(X) = 2$$
$$\to \left\{ \begin{array}{l} 0.9 : \neg on(X, Y) \\ 0.1 : \text{noise} \end{array} \right.$$

This rule's context describes the starting example concisely. *Explain Examples* will consider dropping some of the remaining literals, and thereby generalizing the rule so it applies to examples with different starting states. However, such generalizations do not necessarily improve the score. While they have smaller contexts, they might end up creating more outcomes to describe the new examples, so the penalty term is not guaranteed to improve. The change in the likelihood term will depend on whether the new examples have higher likelihood under the new rule than under the default rule, and on whether the old examples have higher likelihood under the old distribution than under the new one. Quite frequently, the need to cover the new examples will give the new rule a distribution that is closer to random than before, which will usually lead to a decrease in likelihood too large to be overcome by the improvement in the penalty, given the likelihood-penalty trade-off.

Let us assume that, in this case, no predicate can be dropped without worsening the likelihood. The rule will then be integrated into the rule set as is.

- *DropRules* cycles through all the rules in the current rule set, and removes each one in turn from the set. It returns a set of rule sets, each one missing a different rule.

The remaining operators create new rule sets from the input rule set $R$ by repeatedly choosing a rule $r \in R$ and making changes to it to create one or more new rules. These new rules are then integrated into $R$, just as in *ExplainExamples*, to create a new rule set $R'$. Figure 4 shows the the general pseudocode for how this is done. The operators vary in the way they select rules and the changes they make to them. These variations are described for each operator below. (Note that some of the operators, those that deal with deictic references and constants, are only applicable when the action model representation allows these features.)

- *DropLits* selects every rule $r \in R$ $n$ times, where $n$ is the number of literals in the context of $r$; in other words, it selects each $r$ once for each literal in its context. It then creates a new rule $r'$ by removing that literal from $r$'s context; $N$ of Figure 4 is simply the set containing $r'$.

  So, the example *pickup* rule created by *ExplainExamples* would be selected three times, once for *inhand-nil*, once for *clear*($X$), and one for *height*($X$) = 2, and so would create three new rules (each with a different literal missing), three singleton $N$ sets, and three candidate new rule sets $R'$. Since the newly-created $r'$ are generalizations of $r$, they are certain to cover all of $r$'s examples, and so $r$ will be removed from each of the $R'$s.

  The changes suggested by *DropLits* are therefore exactly the same as those suggested by the trimming search in *ExplainExamples*, but there is one crucial difference: *DropLits* attempts to integrate the new rule into the full rule set, instead of just making a quick comparison to the default rule as in Step 2 of *ExplainExamples*. This is because *ExplainExamples* used the trimming search only as a relatively cheap, local heuristic allowing it to decide on a rule size, while *DropLits* uses it to search globally through the space of rule sets, comparing the contributions of various conflicting rules.

- *DropRefs* is an operator used only when deictic references are permitted. It selects each rule $r \in R$ once for each deictic reference in $r$. It then creates a new rule $r'$ by removing that deictic reference from $r$; $N$ is, again, the set containing only $r'$.

  When applying this operator, the *pickup* rule would be selected once, for the reference describing $Y$, and only one new rule set would be returned: one containing the rule without $Y$.

- *GeneralizeEquality* selects each rule $r \in R$ twice for each equality literal in the context to create two new rules: one where the equality is replaced by a $\leq$, and one where it is replaced by a $\geq$. Each rule will then be integrated into the rule set $R$, and the resulting two $R'$s returned. Again, these generalized rules are certain to cover all of $r$'s examples, and so the $R'$s will not contain $r$.

  The context of our *pickup* rule contains one equality literal, *height*($X$) = 1. *GeneralizeEquality* will attempt to replace this literal with *height*($X$) $\leq$ 1 and *height*($X$) $\geq$ 1. In a domain containing more than two blocks, this would be likely to yield interesting generalizations.

- *ChangeRanges* selects each rule $r \in R$ $n$ times for each equality or inequality literal in the context, where $n$ is the total number of values in the range of each literal. Each time it selects $r$ it creates a new rule $r'$ by replacing the numeric value of the chosen (in)equality with another possible value from the range. Note that it is quite possible that some of these new rules will cover no examples, and so will be abandoned. The remaining rules will be integrated into new copies of the rule set as usual.

Thus, if $f()$ ranges over $[1 \ldots n]$, *ChangeRange* would, when applied to a rule containing the inequality $f() < i$, construct rule sets in which $i$ is replaced by all other integers in $[1 \ldots n]$.

Our *pickup* rule contains one equality literal, $height(X) = 1$. In the two-block domain from which our $(s, a, s')$ example was drawn, $height()$ can take on the values 0, 1, and 2, so the rule will, again, be selected thrice, and new rules will be created containing the new equalities. Since the rule constrains $X$ to be *on* something, the new rule containing $height(X) = 0$ can never cover any examples and will certainly be abandoned.

- *SplitOnLits* selects each rule $r \in R$ $n$ times, where $n$ is the number of literals that are absent from the rule's context and deictic references. (The set of absent literals is obtained by applying the available functions and predicates—both primitive and derived—to the terms present in the rule, and removing the literals already present in the rule from the resulting set.) It then constructs a set of new rules. In the case of predicate and inequality literals, it creates one rule in which the positive version of the literal is inserted into the context, and one in which it is the negative version. In the case of equality literals, it constructs a rule for every possible value the equality could take. In either case, rules that cover no examples will be dropped. Any remaining rules corresponding to the one literal are placed in $N$, and they are then integrated into the rule set simultaneously.

  Note that the newly created rules will, between them, cover all the examples that start out covered by the original rule and no others, and that these examples will be split between them.

  The list of literals that may be added to the *pickup* rule consists of $inhand(X)$, $inhand(Y)$, $table(X)$, $table(Y)$, $clear(Y)$, $on(Y, X)$, $on(Y, Y)$, $on(X, X)$, $height(Y) =?$, and all possible applications of *above* and *topstack*. These literals do not make for very interesting examples: adding them to the context will create rules that either cover no examples at all, and so will be abandoned, or that cover the same set of examples as the original rule, and so will be rejected for having the same likelihood but a worse penalty. However, just to illustrate the process, attempting to add in the $height(Y) =?$ predicate will result in the creation of three new rules with $height(Y) = n$ in the context, one for each $n \in [0, 1, 2]$. These rules would be added to the rule set all at once.

- *AddLits* selects each rule $r \in R$ $2n$ times, where $n$ is the number of predicate-based literals that are absent from the rule's context and deictic references, and the 2 reflects the fact that each literal may be considered in its positive or negative form. It constructs a new rule for each literal by inserting that literal into the earliest place in the rule in which its variables are all well-defined: if the literal contains no deictic variables, this will be the context, otherwise this will be the restriction of the last deictic variable mentioned in the literal. (So, if $V_1$ and $V_2$ are deictic variables and $V_1$ appears first, $on(V_1, V_2)$ would be inserted into the restriction of $V_2$.) The resulting rule is then integrated into the rule set.

  The list of literals that may be added to the *pickup* rule is much as for *SplitOnLits*, only without $height(Y) =?$. Again, this process will not lead to anything very interesting in our example, for the same reason. Just as an illustration, $inhand(Y)$ would be chosen twice, once as $\neg inhand(Y)$, and added to the context in each case. Since the context already contains $inhand\text{-}nil$, adding $\neg inhand(Y)$ will be redundant, and adding $inhand(Y)$ will produce a contradiction, so neither rule will be seriously considered.

- *AddRefs* is an operator used only when deictic references are permitted. It selects each rule $r \in R$ $n$ times, where $n$ is the number of literals that can be constructed using the available predicates, the variables in $r$, and a new variable $v$. In each case, it creates a new deictic reference for $v$, using the current literal to define the restriction, and adds this deictic reference to the antecedent of $r$ to construct a new rule, which will then be integrated into the rule set.

  Supposing $V$ is the new variable, the list of literals that would be constructed with the *pickup* rule consists of $inhand(V)$, $clear(V)$, $on(V, X)$, $on(X, V)$, $table(V)$, $on(V, Y)$, $on(Y, V)$, $on(V, V)$, and all possible applications of *above* and *topstack* (which will mirror those for *on*.) They will be used to create deictic references like $V : table(V)$. (A useful reference here, as it allows the rule to describe the "falls on table" outcomes explicitly; this operator is very likely to be accepted at some point in the search.)

- *RaiseConstants* is an operator used only when constants are permitted. It selects each rule $r \in R$ $n$ times, where $n$ is the number of constants among the arguments of $r$'s action. For each constant $c$, it constructs a new rule by creating a new variable and replacing every occurrence of $c$ with it. It then integrates this new rule into the rule set.

- *SplitVariables* is an operator used only when constants are permitted. It selects each rule $r \in R$ $n$ times, where $n$ is the number of variables among the arguments of $r$'s action. For each variable $v$, it goes through the examples covered by the rule $r$ and collects the constants $v$ binds to. Then, it creates a rule for each of these constants by replacing every occurrence of $v$ with that constant. The rules corresponding to one variable $v$ are combined in the set $N$ and integrated into the old rule set together.

We have found that all of these operators are consistently used during learning. While this set of operators is heuristic, it is complete in the sense that every rule set can be constructed from the initial rule set—although, of course, there is no guarantee that the scoring metric will lead the greedy search to the global maximum.

*LearnRules*'s search strategy has one large drawback; the set of learned rules is only guaranteed to be proper on the training set and not on testing data. New test examples could be covered by more than one rule. When this happens, we employ an alternative rule-selection semantics, and return the default rule to model the situation. In this way, we are essentially saying that we don't know what will happen. However, this is not a significant problem; the problematic test examples can always be added to a future training set and used to learn better models. Given a sufficiently large training set, these failures should be rare.

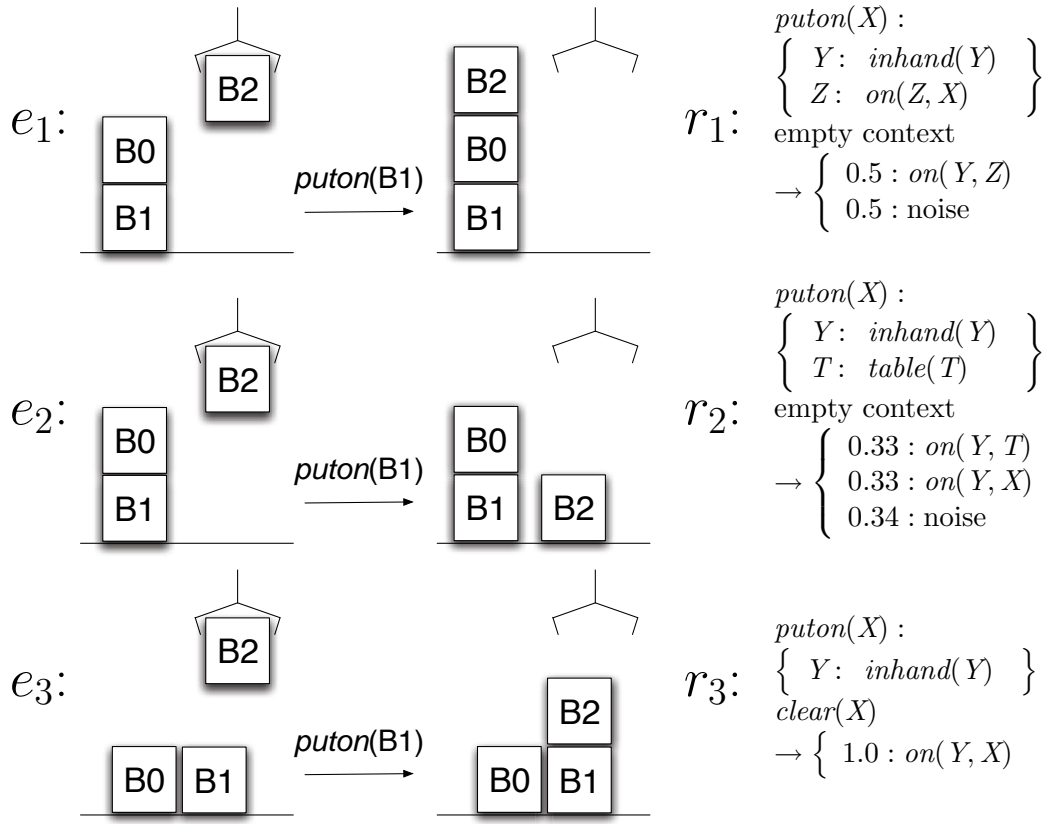

Figure 6: Three training examples in a three blocks world. Each example is paired with an initial rule that *ExplainExamples* might create to model it. In each example, the agent is trying to put block $B2$ onto block $B1$.

## An Example of Rule Set Learning

As an example, consider how *LearnRuleSet* might learn a set of rules to model the three training examples in Figure 6, given the settings of the complexity penalty and noise bound later used in our experiments: $\alpha = 0.5$ and $p_{min} = 0.0000001$. This $p_{min}$ is very low for the three-block domain, since it only has 25 different states, but we use it for consistency.

At initialization, the rule set contains only the default rule; all the changes that occur in these examples are modeled as noise. Since all examples include change, the default rule will have a noise probability of 1.0. We now describe the path the greedy search takes.

During the first round of search the *ExplainExamples* operator suggests adding in new rules to describe these examples. In general, *ExplainExamples* tries to construct rules that are compact, that cover many examples, and that assign a relatively high probability to each covered example. (The latter means that noise outcomes are to be avoided whenever possible.) One reasonable set of rules to be suggested is shown on the right-hand side of Figure 6. Notice that $r_3$ is deterministic, and so high-probability and relatively compact: $e_3$ has a unique initial state, and *ExplainExamples* can take advantage of this. Meanwhile, $e_1$ and $e_2$ have the same starting state, and so the rules explaining them must cover each other's examples. Thus, noise outcomes are unavoidable in both rules, since they lack the necessary deictic references. (Deictic variables are created only to describe the objects whose state changes in the example being explained.)

Now, consider adding one of these rules. There is no guarantee that doing so will constitute an improvement, since a very high complexity penalty $\alpha$ would make any rule look bad, while a high $p_{min}$ would make the default rule look good. To determine what the best move is, the algorithm compares the scores of the rule sets containing each of the proposed rules to the score of the initial rule set containing only the default rule. Let us calculate these scores for the example, starting with the rule set consisting of the rule $r_1$, which covers $e_1$ and $e_2$, and the default rule $r_d$, which covers the remaining example, and which therefore has a noise probability of 1.0. We will use Equation 5, and let a rule's complexity be the number of literals in its body: so, in the case of $r_1$, three. We get:

$$
\begin{aligned}
S(r_1, r_d) &= \sum_{(s,a,s') \in E} \log(\hat{P}(s'|s, a, r_{(s,a)})) - \alpha \sum_{r \in R} PEN(r) \\
&= \log(0.5 + 0.5 * p_{min}) + \log(0.5 * p_{min}) + \log(p_{min}) - \alpha PEN(r_1) - \alpha PEN(r_d) \\
&= \log(0.50000005) + \log(0.00000005) + \log(0.0000001) - 0.5 * 3 - 0.5 * 0 \\
&= -0.301 - 7.301 - 7 - 1.5 \\
&= -16.101
\end{aligned}
$$

So, the rule set containing $r_1$ has a score of $-16.101$. Similar calculations show that the rule sets containing $r_2$ and $r_3$ have scores of $-10.443$ and $-15.5$ respectively. Since the initial rule set has a score of $-21$, all of these new rule sets are improvements, but the one containing $r_2$ is best, and will be picked by the greedy search. The new rule set is now:

$$
puton(X) : \left\{ \; Y : inhand(Y), \, T : table(T) \; \right\}
$$

empty context

$$
\rightarrow \left\{
\begin{array}{l}
0.33 : on(Y, T) \\
0.33 : on(Y, X) \\
0.34 : noise
\end{array}
\right.
$$

default rule:

$$
\rightarrow \left\{
\begin{array}{l}
1.0 : no \; change \\
0.0 : noise
\end{array}
\right.
$$

Notice that all the training examples are covered by a non-default rule. In this situation, the default rule does not cover any examples and has no probability assigned to the noise outcome.

At the next step, the search has to decide between altering an existing rule, and introducing another rule to describe an example currently covered by the default rule. Since the default rule covers no examples, altering the single rule in the rule set is the only option. The operators most likely to score highly are those that can get rid of that noise outcome, which is there because the rule has no means of referring to the block above $X$ in $e_1$. The appropriate operator is therefore *AddRefs*, which can introduce a new deictic reference describing that block. Of course, this increases the size of the rule, and so its complexity, and in addition it means that the rule no longer applies to $e_3$, leaving that example to be handled by the default rule. However, the new rule set raises the probabilities of all the examples enough to compensate for the increase in complexity, and so it ends up with a score of $-10.102$, which is a clear improvement on $-10.443$. This is the highest score obtainable at this step, so the algorithm alters the rule set to get:

$$puton(X) : \{\ Y : inhand(Y), T : table(T), Z : on(Z, X)\ \}$$
empty context
$$\rightarrow \left\{ \begin{array}{l} 0.5 : on(Y, Z) \\ 0.5 : on(Y, T) \end{array} \right.$$
default rule:
$$\rightarrow \left\{ \begin{array}{l} 0.0 : \text{no change} \\ 1.0 : \text{noise} \end{array} \right.$$

Now that the default rule covers $e_3$, *ExplainExamples* has something to work with again. Adding in $r_3$ will get rid of all noise, and yield a much improved score of $-4.602$. Again, this is the biggest improvement that can be made, and the rule set becomes:

$$puton(X) : \{\ Y : inhand(Y), T : table(T), Z : on(Z, X)\ \}$$
empty context
$$\rightarrow \left\{ \begin{array}{l} 0.5 : on(Y, Z) \\ 0.5 : on(Y, T) \end{array} \right.$$
$$puton(X) : \{\ Y : inhand(Y)\ \}$$
$$clear(X)$$
$$\rightarrow \{\ 1.0 : on(Y, X)\ \}$$
default rule:
$$\rightarrow \left\{ \begin{array}{l} 1.0 : \text{no change} \\ 0.0 : \text{noise} \end{array} \right.$$

Note that this rule could not have been added earlier because $e_3$ was also covered by the first rule added, $r_2$, before it was specialized. Thus, adding $r_3$ to the rule set containing $r_2$ would have knocked $r_2$ out, and caused examples $e_1$ and $e_2$ to be explained as noise by the default rule, which would have reduced the overall score. (It is, however, possible for a rule to knock out another and yet improve the score: it just requires a more complicated set of examples.)

Learning continues with more search. Attempts to apply the rule-altering operators to the current rules will either make them bigger without changing the likelihood, or will lead to the creation of some noise outcomes. Dropping either rule will add noise probability to the default rule and lower the score. Since there are no extra examples to be explained, no operator can improve the score, and the search stops at this rule set. It seems like a reasonable rule set for this domain: one rule covers what happens when we try to *puton* a clear block, and one describes when we try to *puton* a block that has another block *on* it. Ideally, we would like the first rule to generalize to blocks that have something *above* them, instead of just *on*, but to notice that we would need examples containing higher stacks.

### 5.1.6 DIFFERENT VERSIONS OF THE ALGORITHM

By making small variations in the *LearnRuleSet* algorithm, we can learn different types of rule sets. This will be important for evaluating the algorithm.

To explore the effects of constants in the rules, we will evaluate three different versions of rule learning: *propositional*, *relational*, and *deictic*. For propositional rule learning, *ExplainExamples* creates initial trimmed rules with constants but never introduces variables. None of the search operators that introduce variables are used. Thus, the learned rules are guaranteed to be propositional—they cannot generalize across the identities of specific objects. For relational rule learning, variables are allowed in rule action arguments but the search operators are not allowed to introduce deictic references. *ExplainExamples* creates rules with constants that name objects, as long as those constants do not already have a variable in the action argument list mapped to them. Finally, for deictic rule learning, no constants are allowed. We will see that deictic learning provides a strong bias that can improve generalization.

To demonstrate that the addition of noise and deictic references can result in better rules, we will learn action models that do not have these enhancements. Again, this can be done by changing the algorithm in minor ways. To disallow noise, we set the rule noise probability to zero, which means that we must then constrain outcome sets to contain an outcome for every example where change was observed; rules that cannot express all the changes are abandoned. To disallow deictic references, we disable the operators that introduce them, and have *ExplainExamples* create an empty deictic reference set.

## 5.2 Learning Concepts

The contexts and deictic references of NDRs can make use of concept predicates and functions as well as primitive ones. These concepts can be specified by hand, or learned using a rather simple algorithm, *LearnConcepts*, which uses *LearnRuleSet* as a subprocedure for testing concept usefulness. The algorithm works by constructing increasingly complex concepts, and then running *LearnRuleSet* and checking what concepts appear in the learned rules. The first set is created by applying the operators in Figure 7 to literals built with the original language. Subsequent sets of concepts are constructed using the literals that proved useful on the latest run; concepts that have been tried before, or that are always true or always false across all examples, are discarded. The search ends when none of the new concepts prove useful.

As an example, consider the predicate *topstack* in a simple blocks world, which could be discovered as follows. In the first round of learning, the literal $on(X_1, X_2)$ is used to define the new predicate $n(Y_1, Y_2) := on^*(Y_1, Y_2)$, which is true when $Y_1$ is stacked above $Y_2$. Assuming this new predicate appears in the learned rules, it can then be used in the second round of learning, to define, among others, $m(Z_1, Z_2) := n(Z_1, Z_2) \land clear(Z_1)$. By ensuring that $Z_1$ is *clear*, this predicate will be true only when $Z_1$ is the highest block in the stack containing $Z_2$. This notion of *topstack* can be used to determining what will happen with the gripper tries to pick up $Z_2$. Because it descends from above, it will likely grasp the block on the top of the stack instead.

Since our concept language is quite rich, overfitting (e.g., by learning concepts that can be used to identify individual examples) can be a serious problem. We handle this in the

$$
\begin{aligned}
p(X) &\rightarrow n := QY.p(Y) \\
p(X_1, X_2) &\rightarrow n(Y_2) := QY_1.p(Y_1, Y_2) \\
p(X_1, X_2) &\rightarrow n(Y_1) := QY_2.p(Y_1, Y_2) \\
p(X_1, X_2) &\rightarrow n(Y_1, Y_2) := p^*(Y_1, Y_2) \\
p(X_1, X_2) &\rightarrow n(Y_1, Y_2) := p^+(Y_1, Y_2) \\
p_1(X_1), p_2(X_2) &\rightarrow n(Y_1) := p_1(Y_1) \wedge p_2(Y_1) \\
p_1(X_1), p_2(X_2, X_3) &\rightarrow n(Y_1, Y_2) := p_1(Y_1) \wedge p_2(Y_1, Y_2) \\
p_1(X_1), p_2(X_2, X_3) &\rightarrow n(Y_1, Y_2) := p_1(Y_1) \wedge p_2(Y_2, Y_1) \\
p_1(X_1, X_2), p_2(X_3, X_4) &\rightarrow n(Y_1, Y_2) := p_1(Y_1, Y_2) \wedge p_2(Y_1, Y_2) \\
p_1(X_1, X_2), p_2(X_3, X_4) &\rightarrow n(Y_1, Y_2) := p_1(Y_1, Y_2) \wedge p_2(Y_2, Y_1) \\
p_1(X_1, X_2), p_2(X_3, X_4) &\rightarrow n(Y_1, Y_2) := p_1(Y_1, Y_2) \wedge p_2(Y_1, Y_1) \\
p_1(X_1, X_2), p_2(X_3, X_4) &\rightarrow n(Y_1, Y_2) := p_1(Y_1, Y_2) \wedge p_2(Y_2, Y_2) \\
f(X) = c &\rightarrow n() := \#Y.f(Y) = c \\
f(X) \leq c &\rightarrow n() := \#Y.f(Y) \leq c \\
f(X) \geq c &\rightarrow n() := \#Y.f(Y) \geq c
\end{aligned}
$$

Figure 7: Operators used to invent a new predicate $n$. Each operator takes as input one or more literals, listed on the left. The $p$s represent old predicates; $f$ represents an old function; $Q$ can refer to $\forall$ or $\exists$; and $c$ is a numerical constant. Each operator takes a literal and returns a concept definition. These operators are applied to all of the literals used in rules in a rule set to create new predicates.

expected way: by introducing a penalty term, $\alpha' c(R)$, to create a new scoring metric

$$
S'(R) = S(R) - \alpha' c(R)
$$

where $c(R)$ is the number of distinct concepts used in the rule set $R$ and $\alpha'$ is a scaling parameter. This new metric $S'$ is now used by *LearnRuleSet*; it avoids overfitting by favoring rule sets that use fewer derived predicates. (Note that the fact that $S'$ cannot be factored by rule, as $S$ was, does not matter, since the factoring was used only by *InduceOutcomes* and *LearnParameters*, neither of which can change the number of concepts used in the relevant rule: outcomes contain only primitive predicates.)

## 5.3 Discussion

The rule set learning challenge addressed in this section is complicated by the need to learn the structure of the rules, the numeric parameters associated with the outcome distributions, and the definitions of derived predicates for the modeling language. The *LearnConcepts* algorithm is conceptually simple, and performs this simultaneous learning effectively, as we will see in the experiments in Section 7.2.

The large number of possible search operators might cause concern about the overall computational complexity of the *LearnRuleSet* algorithm. Although this algorithm is expensive, the set of search operators were designed to control this complexity by attempting keep the number of rules in the current set as small as possible.

At each step of search, the number of new rule sets that are considered depends on the current set of rules. The *ExplainExamples* operator creates $d$ new rule sets, where $d$ is the number of examples covered by the default rule. Since the search starts with a rule set containing only the default rule, $d$ is initially equal to the number of training examples. However, *ExplainExamples* was designed to introduce rules that cover many examples, and in practice $d$ grows small quickly. All of the other operators can create $O(rm)$ new rule sets, where $r$ is the number of rules in the current set and $m$ depends on the specific operator. For example, $m$ could be the number of literals that can be dropped from the context of a rule by the *DropLits* operator. Although $m$ can be large, $r$ stays small in practice because the search starts with only the default rule and the complexity penalty favors small rule sets.

Because we ensure that the score increases at each search step, the algorithm is guaranteed to converge to a (usually local) optimum. There is not, however, any guarantee about how quickly it will get there. In practice, we found that the algorithm converged quickly in the test domains. The *LearnRuleSet* algorithm never took more that 50 steps and the *LearnConcepts* outer loop never cycled more than 5 times. The entire algorithm never took more than six hours to run on a single processor, although significant effort was made to cache intermediate computations in the final implementation.

In spite of this, we realize that, as we scale up to more complex domains, this approach will eventually become prohibitively expensive. We plan to handle this problem by developing new algorithms that learn concepts, rules, and rule parameters in an online manner, with more directed search operators. However, we leave this more complex approach to future work.

## 6. Planning

Some of the experiments in Section 7.2 involve learning models of complex actions where true models of the dynamics, at the level of relational rules, are not available for evaluation. Instead, the learned models are evaluated by planning and executing actions. There are many possible ways to plan. In this work, we explore MDP planning.

A MDP (Puterman, 1999) is a 4-tuple $(\mathcal{S}, \mathcal{A}, T, \mathcal{R})$. $\mathcal{S}$ is the set of possible states, $\mathcal{A}$ is a set of possible actions, and $T$ is a distribution that encodes the transition dynamics of the world, $T(s'|s,a)$. Finally, $\mathcal{R}$ is a reward signal that maps every state to a real value. A policy or plan $\pi$ is a (possibly stochastic) mapping from states to actions. The expected amount of reward that can be achieved by executing $\pi$ starting in $s$ is called the *value* of $s$ and defined as $V_\pi(s) = \mathbf{E}[\sum_{i=0}^{\infty} \gamma^i \mathcal{R}(s_i)|\pi]$, where the $s_i$ are the states that can be reached by $\pi$ at time $i$. The discount factor $0 \le \gamma < 1$ favors more immediate rewards. The goal of MDP planning is to find the policy $\pi^*$ that will achieve the most reward over time. This optimal policy can be found by solving the set of Bellman equations,

$$\forall s, V_\pi(s) = \mathcal{R}(s) + \gamma \sum_{s' \in \mathcal{S}} T(s'|s, \pi(a)) V_\pi(s'). \tag{8}$$

In our application, the action set $\mathcal{A}$ and the state set $\mathcal{S}$ are defined by the world we are modeling. A rule set $R$ defines the transition model $T$ and the reward function $\mathcal{R}$ is defined by hand.

Because we will be planning in large domains, it will be difficult to solve the Bellman equations exactly. As an approximation, we implemented a simple planner based on the sparse sampling algorithm (Kearns, Mansour, & Ng, 2002). Given a state $s$, it creates a tree of states (of predefined depth and branching factor) by sampling forward using a transition model, computes the value of each node using the Bellman equation, and selects the action that has the highest value.

We adapt the algorithm to handle noisy outcomes, which do not predict the next state, by estimating the value of the unknown next state as a fraction of the value of staying in the same state: i.e., we sample forward as if we had stayed in the same state and then scale down the value we obtain. Our scaling factor was 0.75, and our depth and branching factor were both four.

This scaling method is only a guess at what the value of the unknown next state might be; because noisy rules are partial models, there is no way to compute the value explicitly. In the future, we would like to explore methods that learn to associate values with noise outcomes. For example, the value of the outcome where a tower of blocks falls over is different if the goal is to build a tall stack of blocks than if the goal is to put all of the blocks on the table.

While this algorithm will not solve hard combinatorial planning problems, it will allow us to choose actions that maximize relatively simple reward functions. As we will see in the next section, this is enough to distinguish good models from poor ones. Moreover, the development of first-order planning techniques is an active field of research (AIPS, 2006).

## 7. Evaluation

In this section, we demonstrate that the rule learning algorithm is robust on a variety of low-noise domains, and then show that it works in our intrinsically noisy simulated blocks world domain. We begin by describing our test domains, and then report a series of experiments.

### 7.1 Domains

The experiments we performed involve learning rules for the domains which are briefly described in the following sections.

#### 7.1.1 SLIPPERY GRIPPER

The slippery gripper domain, inspired by the work of Draper et al. (1994), is an abstract, symbolic blocks world with a simulated robotic arm, which can be used to move the blocks around on a table, and a nozzle, which can be used to paint the blocks. Painting a block might cause the gripper to become wet, which makes it more likely that it will fail to manipulate the blocks successfully; fortunately, a wet gripper can be dried.

$$pickup(X,Y) : on(X,Y), clear(X), \atop inhand\text{-}nil, block(X), block(Y), \neg wet, \rightarrow \left\{ \begin{array}{l} .7 : \begin{array}{l} inhand(X), \neg clear(X), \neg inhand\text{-}nil, \\ \neg on(X,Y), clear(Y) \end{array} \\ .2 : on(X, \text{TABLE}), \neg on(X,Y) \\ .1 : \text{no change} \end{array} \right.$$

$$pickup(X,Y) : on(X,Y), clear(X), \atop inhand\text{-}nil, block(X), block(Y), wet \rightarrow \left\{ \begin{array}{l} .33 : \begin{array}{l} inhand(X), \neg clear(X), \neg inhand\text{-}nil, \\ \neg on(X,Y), clear(Y) \end{array} \\ .33 : on(X, \text{TABLE}), \neg on(X,Y) \\ .34 : \text{no change} \end{array} \right.$$

$$pickup(X,Y) : on(X,Y), clear(X), \atop inhand\text{-}nil, block(X), table(Y), wet \rightarrow \left\{ \begin{array}{l} .5 : \begin{array}{l} inhand(X), \neg clear(X), \neg inhand\text{-}nil, \\ \neg on(X,Y) \end{array} \\ .5 : \text{no change} \end{array} \right.$$

$$pickup(X,Y) : on(X,Y), clear(X), \atop inhand\text{-}nil, block(X), table(Y), \neg wet \rightarrow \left\{ \begin{array}{l} .8 : \begin{array}{l} inhand(X), \neg clear(X), \neg inhand\text{-}nil, \\ \neg on(X,Y) \end{array} \\ .2 : \text{no change} \end{array} \right.$$

$$puton(X,Y) : clear(Y), inhand(X), \atop block(Y) \rightarrow \left\{ \begin{array}{l} .7 : \begin{array}{l} inhand\text{-}nil, \neg clear(Y), \neg inhand(X), \\ on(X,Y), clear(X) \end{array} \\ .2 : \begin{array}{l} on(X, \text{TABLE}), clear(X), inhand\text{-}nil, \\ \neg inhand(X) \end{array} \\ .1 : \text{no change} \end{array} \right.$$

$$puton(X, \text{TABLE}) : inhand(X) \rightarrow \left\{ \begin{array}{l} .8 : \begin{array}{l} on(X, \text{TABLE}), clear(X), inhand\text{-}nil, \\ \neg inhand(X) \end{array} \\ .2 : \text{no change} \end{array} \right.$$

$$paint(X) : block(X) \rightarrow \left\{ \begin{array}{l} .6 : painted(X) \\ .1 : painted(X), wet \\ .3 : \text{no change} \end{array} \right.$$

$$dry : \text{no context} \rightarrow \left\{ \begin{array}{l} .9 : \neg wet \\ .1 : \text{no change} \end{array} \right.$$

Figure 8: Eight relational planning rules that model the slippery gripper domain.

Figure 8 shows the set of rules that model this domain. Individual states represent world objects as intrinsic constants and experimental data is generated by sampling from the rules. In Section 7.2.1, we will explore how the learning algorithms of Section 5 compare as the number of training examples is scaled in a single complex world.

### 7.1.2 Trucks and Drivers

Trucks and drivers is a logistics domain, adapted from the 2002 AIPS international planning competition (AIPS, 2002). There are four types of constants: trucks, drivers, locations, and objects. Trucks, drivers and objects can all be at any of the locations. The locations are connected with paths and links. Drivers can board trucks, exit trucks, and drive trucks between locations that are linked. Drivers can also walk, without a truck, between locations that are connected by paths. Finally, objects can be loaded and unloaded from trucks.

A set of rules is shown in Figure 9. Most of the actions are simple rules which succeed or fail to change the world. However, the walk action has an interesting twist. When drivers try to walk from one location to another, they succeed most of the time, but some of the

$$load(O, T, L) : \\ at(T, L), at(O, L) \quad \rightarrow \quad \begin{cases} .9 : \neg at(O, L), in(O, T) \\ .1 : \text{no change} \end{cases}$$

$$unload(O, T, L) : \\ in(O, T), at(T, L) \quad \rightarrow \quad \begin{cases} .9 : at(O, L), \neg in(O, T) \\ .1 : \text{no change} \end{cases}$$

$$board(D, T, L) : \\ at(T, L), at(D, L), empty(T) \quad \rightarrow \quad \begin{cases} .9 : \neg at(D, L), driving(D, T), \neg empty(T) \\ .1 : \text{no change} \end{cases}$$

$$disembark(D, T, L) : \\ at(T, L), driving(D, T) \quad \rightarrow \quad \begin{cases} .9 : \neg driving(D, T), at(D, L), empty(T) \\ .1 : \text{no change} \end{cases}$$

$$drive(T, FL, TL, D) : \\ driving(D, T), at(T, FL), link(FL, TL) \quad \rightarrow \quad \begin{cases} .9 : at(T, TL), \neg at(T, FL) \\ .1 : \text{no change} \end{cases}$$

$$walk(D, FL, TL) : \\ at(D, FL), path(FL, TL) \quad \rightarrow \quad \begin{cases} .9 : at(D, TL), \neg at(D, FL) \\ .1 : \begin{array}{l} \text{pick } X \text{ s.t. } path(FL, X) \\ at(D, X), \neg at(D, FL) \end{array} \end{cases}$$

Figure 9: Six rules that encode the world dynamics for the trucks and drivers domain.

time they arrive at a randomly chosen location that is connected by some path to their origin location.

The representation presented here cannot encode this action efficiently. The best rule set has a rule for each origin location, with outcomes for every location that the origin is linked to. Extending the representation to allow actions like walk to be represented as a single rule is an interesting area for future work.

Like in the slippery gripper domain, individual states represent world objects as intrinsic constants and experimental data is generated by sampling from the rules. The trucks and drivers dynamics is more difficult to learn but, as we will see in Section 7.2.1, can be learned with enough training data.

### 7.1.3 Simulated Blocks World

To validate the rule extensions in this paper, Section 7.2 presents experiments in a rigid body, simulated physics blocks world. This section describes the logical interface to the simulated world. A description of the extra complexities inherent in learning the dynamics of this world was presented in Section 1.

We now define the interface between the symbolic representation that we use to describe action dynamics and a physical domain such as the simulated blocks world. The perceptual system produces states that contain skolem constants. The logical language includes the binary predicate $on(X, Y)$, which is defined as '$X$ exerts a downward force on $Y$' and obtained by querying the internal state of the simulator, and unary typing predicates *table* and *block*. The actuation system translates actions into sequences of motor commands in the simulator. Actions always execute, regardless of the state of the world. We define two actions; both have parameters that allow the agent to specify which objects it intends to manipulate. The $pickup(X)$ action centers the gripper above $X$, lowers it until it hits something, grasps, and raises the gripper. Analogously, the $puton(X)$ action centers the gripper above $X$, lowers until it encounters pressure, opens it, and raises it.

By using a simulator we are sidestepping the difficult "pixels-to-predicates" problem that occurs whenever an agent has to map domain observations into an internal representation. Primitive predicates defined in terms of the internal state of the simulation are simpler and cleaner than observations of the real world would be. They also make the domain completely observable: a prerequisite for our learning and planning algorithms. Choosing the set of predicates to observe is important. It can make the rule learning problem very easy or very hard, and the difficulty of making this choice is magnified in richer settings. The limited language described above balances these extremes by providing *on*, which would be difficult to derive by other means, but not providing predicates such as *inhand* and *clear*, that can be learned.

## 7.2 Experiments

This section describes two sets of experiments. First, we compare the learning of deictic, relational, and propositional rules on the slippery gripper and trucks and drivers data. These domains are modeled by planning rules, contain intrinsic constants, and are not noisy, and thus allow us to explore the effect of deictic references and constants in the rules directly. Then, we describe a set of experiments that learns rules to model data from the simulated blocks world. This data is inherently noisy and contains skolem constants. As a result, we focus on evaluating the full algorithm by performing ablation studies that demonstrate that deictic references, noise outcomes, and concepts are all required for effective learning.

All of the experiments use examples, $(s, a, s') \in \mathbf{E}$, generated by randomly constructing a state $s$, randomly picking the arguments of the action $a$, and then executing the action in the state to generate $s'$. The distribution used to construct $s$ is biased to guarantee that, in approximately half of the examples, $a$ has a chance to change the state. This method of data generation is designed to ensure that the learning algorithms will always have data which is representative of the entire model that they should learn. Thus, these experiments ignore the problems an agent would face if it had to generate data by exploring the world.

### 7.2.1 Learning Rule Sets with No Noise

When we know the model used to generate the data, we can evaluate our model with respect to a set of similarly generated test examples $\mathbf{E}$ by calculating the average *variational distance* between the true model $P$ and the estimate $\hat{P}$,

$$VD(P, \hat{P}) = \frac{1}{|\mathbf{E}|} \sum_{E \in \mathbf{E}} |P(E) - \hat{P}(E)| \ .$$

Variational distance is a suitable measure because it clearly favors similar distributions, and yet is well-defined when a zero probability event is observed. (As can happen when a non-noisy rule is learned from sparse data and does not have as many outcomes as it should.)

These comparisons are performed for four actions. The first two, paint and pickup, are from the slippery gripper domain, while the second two, drive and walk, are from the trucks and drivers domain. Each action presents different challenges for learning. *Paint* is a simple action where more than one outcome can lead to the same successor state (as described in Section 4.1). *Pickup* is a complex action that must be represented by more than one planning rule. *Drive* is a simple action that has four arguments. Finally, *walk* is a complicated action uses the path connectivity of the world in its noise model for lost pedestrians. The slippery gripper actions were performed in a world with four blocks. The trucks and driver actions were performed in a world with two trucks, two drivers, two objects, and four locations.

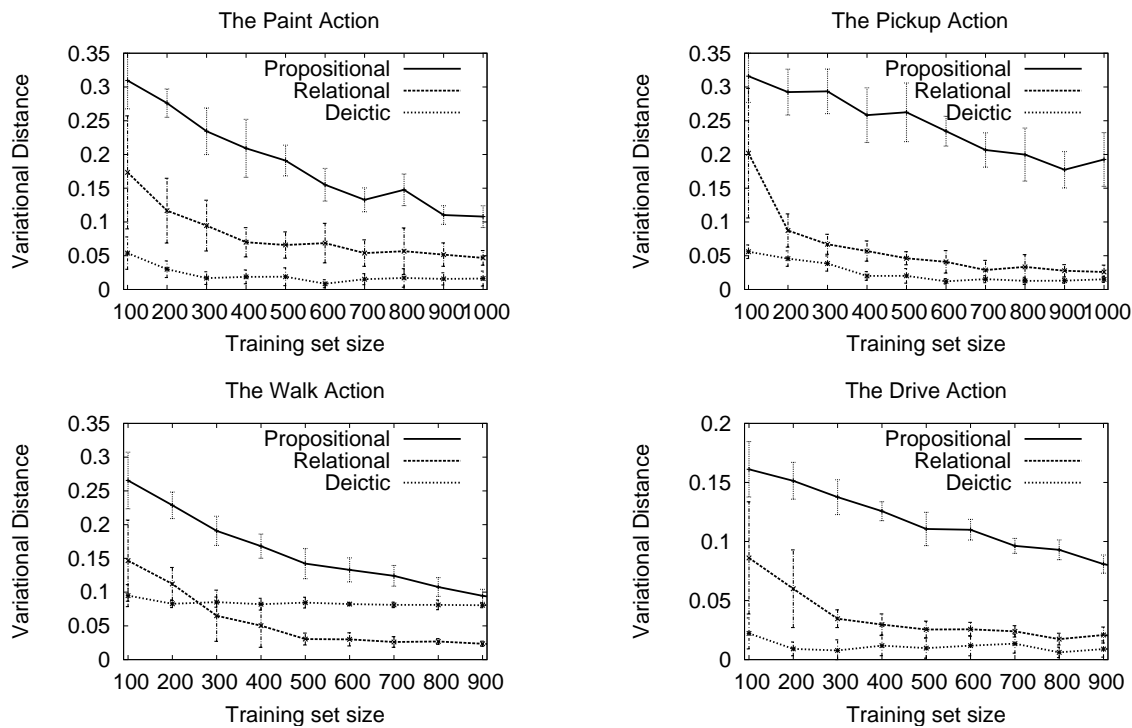

Figure 10: Variational distance as a function of the number of training examples for propositional, relational, and deictic rules. The results are averaged over ten trials of the experiment. The test set size was 400 examples.

We compare three versions of the algorithm: deictic, which includes the full rules language and does not allow constants; relational, which allows variables and constants but no deictic references; and propositional, which has constants but no variables. Figure 10 shows the results. The relational learning consistently outperforms propositional learning; this implies that the variable abstractions are useful. In all cases except for the *walk* action, the deictic learner outperforms the relational learner. This result implies that forcing the rules to only contain variables is preventing overfitting and learning better models. The results on the *walk* action are more interesting. Here, the deictic learner cannot actually represent the optimal rule; it requires a noise model that is too complex. The deictic learner quickly learns the best rule it can, but the relational and propositional learners eventually learn better rule sets because they can use constants to more accurately model the walkers moving to random locations.

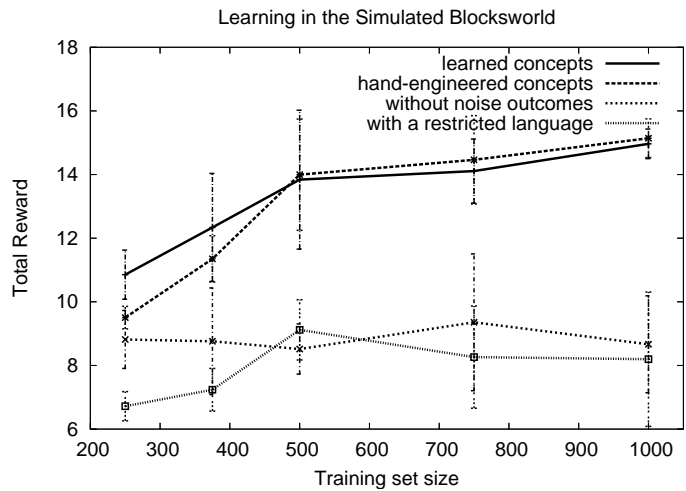

Figure 11: The performance of various action model variants as a function of the number of training examples. All data points were averaged over five planning trials for each of the three rule sets learned from different training data sets. For comparison, the average reward for performing no actions is 9.2, and the reward obtained when a human directed the gripper averaged 16.2.

In these experiments, we see that variable abstraction helps to learn from less data, and that deictic rules, which abstract the most aggressively, perform the best, as long as they can represent the model to be learned. In the next section, we will only consider deictic rules, since we will be working in a domain with simulated perception that does not have access to objects' identities and names them using skolem constants.

### 7.2.2 Learning in the Blocks World Simulator

Our final experiment demonstrates that both noise outcomes and complicated concepts are necessary to learn good action models for the blocks world simulator.

When the true model is not known, we evaluate the learned model by using it to plan and estimating the average reward it gets. The reward function we used in simulated blocks world was the average height of the blocks in the world, and the breadth and depth of the search for the sampling planner were both four. During learning, we set $\alpha$ to 0.5 and $p_{min}$ to 0.0000001.

We tested four action model variants, varying the training set size; the results are shown in Figure 11. The curve labeled 'learned concepts' represents the full algorithm as presented in this paper. Its performance approaches that obtained by a human expert, and is comparable to that of the algorithm labeled 'hand-engineered concepts' that did not do concept learning, but was, instead, provided with hand-coded versions of the concepts *clear*, *inhand*, *inhand-nil*, *above*, *topstack*, and *height*. The concept learner discovered all of these, as well as other useful predicates, e.g., $p(X, Y) := clear(Y) \wedge on(Y, X)$, which we will call *onclear*. This could be why its action models outperformed the hand-engineered ones slightly on small training sets. In domains less well-studied than the blocks world, it might be less obvious what the useful concepts are; the concept-discovery technique presented here should prove helpful.

The remaining two model variants obtained rewards comparable to the reward for doing nothing at all. (The planner did attempt to act during these experiments, it just did a poor job.) In one variant, we used the same full set of predefined concepts but the rules could not have noise outcomes. The requirement that they explain every action effect led to significant overfitting and a decrease in performance. The other rule set was given the traditional blocks world language, which does not include *above*, *topstack*, or *height*, and allowed to learn rules with noise outcomes. We also tried a full-language variant where noise outcomes were allowed, but deictic references were not: the resulting rule sets contained only a few very noisy rules, and the planner did not attempt to act at all. The poor performance of these ablated versions of our representation shows that all three of our extensions are essential for modeling the simulated blocks world domain.

A human agent commanding the gripper to solve the same problem received an average total reward of 16.2, which was below the theoretical maximum due to unexpected action outcomes. Thus, the ND rules are performing at near-human levels, suggesting that this representation is a reasonable one for this problem. It also suggests that our planning approximations and learning bounds are not limiting performance. Traditional rules, which face the challenge of modeling all the transitions seen in the data, have a much larger hypothesis space to consider while learning; it is not surprising that they generalize poorly and are consistently out-performed by the NDRs.

Informally, we can also report that NDR algorithms execute significantly faster than the traditional ones. On one standard desktop PC, learning NDRs takes minutes while learning traditional rules can take hours. Because noisy deictic action models are generally more compact than traditional ones (they contain fewer rules with fewer outcomes) planning is much faster as well.

To get a better feel for the types of rules learned, here are two interesting rules produced by the full algorithm.

$$pickup(X) : \left\{ \begin{array}{l} Y : onclear(X, Y), \ Z : on(Y, Z), \\ T : table(T) \end{array} \right\}$$

$$inhand\text{-}nil, \, size(X) < 2$$

$$\rightarrow \left\{ \begin{array}{l} .80 : \ \neg on(Y, Z) \\ .10 : \ \neg on(X, Y) \\ .10 : \ \neg on(X, Y), on(Y, T), \neg on(Y, Z) \end{array} \right.$$

This rule applies when the empty gripper is asked to pick up a small block $X$ that sits on top of another block $Y$. The gripper grabs both with a high probability.

$$puton(X) : \left\{ \begin{array}{l} Y : topstack(Y, X), \ Z : inhand(Z), \\ T : table(T) \end{array} \right\}$$
$$size(Y) < 2$$
$$\rightarrow \left\{ \begin{array}{ll} .62 : & on(Z, Y) \\ .12 : & on(Z, T) \\ .04 : & on(Z, T), on(Y, T), \neg on(Y, X) \\ .22 : & \text{noise} \end{array} \right.$$

This rule applies when the gripper is asked to put its contents, $Z$, on a block $X$ which is inside a stack topped by a small block $Y$. Because placing things on a small block is chancy, there is a reasonable probability that $Z$ will fall to the table, and a small probability that $Y$ will follow.

## 8. Discussion

In this paper, we developed a probabilistic action model representation that is rich enough to be used to learn models for planning in a physically simulated blocks world. This is a first step towards defining representations and algorithms that will enable learning in more complex worlds.

### 8.1 Related Work

The problem of learning deterministic action models is well studied. Most work in this area (Shen & Simon, 1989; Gil, 1993, 1994; Wang, 1995) has focused on incrementally learning planning operators by interacting with simulated worlds. However, all of this work assumes that the learned models are completely deterministic.

Oates and Cohen (1996) did the earliest work on learning probabilistic planning operators. Their rules are factored and can apply in parallel. However, their representation is strictly propositional, and allows each rule to contain only a single outcome. In our previous work, we developed algorithms for learning probabilistic relational planning operators (Pasula, Zettlemoyer, & Kaelbling, 2004). Unfortunately, neither of these probabilistic algorithms are robust enough to learn in complex, noisy environments like the simulated blocks world.

One previous system that comes close to this goal is the TRAIL learner (Benson, 1996). TRAIL learns an extended version of Horn clauses in noisy environments by applying inductive logic programming (ILP) learning techniques that are robust to noise. TRAIL introduced deictic references that name objects based on their functional relationships to arguments of the actions. Our deictic references, with their exists-unique quantification semantics, are a generalization of Benson's original work. Moreover, TRAIL models continuous actions and real-valued fluents, which allows it to represent some of the most complex models to date, including the knowledge required to pilot a realistic flight simulator. However, the rules that TRAIL learns are in a limited probabilistic representation that can not represent all possible transition distributions. TRAIL also does not include any mechanisms for learning new predicates.

All of this work on action model learning has used different versions of greedy search for rule structure learning, which is closely related to and inspired by the learning with version spaces of Mitchell (1982) and later ILP work (Lavrač & Džeroski, 1994). In this paper, we also explore, for the first time, a new way of moving through the space of rule sets by using the noise rule as an initial rule set. We have found that this approach works well in practice, avoiding the need for a hand-selected initial rule set and allowing our algorithm to learn in significantly more complex environments.

As far as we know, no work on learning action models has explored learning concepts. In the ILP literature, recent work (Assche, Vens, Blockeel, & Džeroski, 2004) has shown that adding concept learning to decision tree learning algorithms improves classification performance.

Outside of action learning, there exists much related research on learning probabilistic models with relational or logical structure. A complete discussion is beyond the scope of this paper, but we present a few highlights. Some work learns representations that are relational extension of Bayesian networks. For a comprehensive example, see work by Getoor (2001). Other work extends research in ILP by incorporating probabilistic dependencies. For example, see the wide range of techniques presented by Kersting (2006). Additionally, there is recent work on learning Markov logic networks (Richardson & Domingos, 2006; Kok & Domingos, 2005), which are log-linear models with features that are defined by first-order logical formulae. The action models and action model learning algorithms in this paper are designed to represent action effects, a special case of the more general approaches listed above. As we have discussed in Section 2, by tailoring the representation to match the model to be learnt, we simplify learning.

Finally, let us consider work related to the NDR action model representation. The most relevant approach is PPDDL, a representation language for probabilistic planning operators and problem domains (Younes & Littman, 2004). The NDR representation was partially inspired by PPDDL operators but includes restrictions to make it easier to learn and extensions, such as noise outcomes, that are required to effectively model the simulated blocks world. In the future, the algorithms in this paper could be extended to learn full PPDDL rules. Also, PPDDL planning algorithms (for examples, see the papers from recent planning competitions) could be adapted to improve the simple planning presented in Section 6. In a more general sense, NDRs are related to all the other probabilistic relational representations that are designed to model dependencies across time. For examples, see work on relational dynamic Bayesian networks (Sanghai, Domingos, & Weld, 2005), which are a specialization of PRMs, and logical hidden Markov models (Kersting, Raedt, & Raiko, 2006), which come from the ILP research tradition. These approaches make a different set of modeling assumptions that are not as closely tied to the planning representations that NDR models extend.

## 8.2 Future and Ongoing Work

There remains much to be done in the context of learning probabilistic planning rules. First of all, it is very likely that when this work is applied to additional domains (such as more realistic robotic applications or dialogue systems) the representation will need to be adapted, and the search operators adjusted accordingly. Some possible changes mentioned in this article include allowing the rules to apply in parallel, so different rules could apply to different aspects of the state, and extending the outcomes to include quantifiers, so actions like *walk*, from the trucks and drivers domain in Section 7.1.2, could be described using a single rule. A more significant change we intend to pursue is expanding this approach to handle partial observability, possibly by incorporating some of the techniques from work on deterministic learning (Amir, 2005). We also hope to make some changes that will make using the rules easier, such as associating values with the noise outcomes to help a planner decide whether they should be avoided.

A second research direction involves the development of new algorithms that learn probabilistic operators in an incremental, online manner, similar to the learning setup in the deterministic case (Shen & Simon, 1989; Gil, 1994; Wang, 1995). This has the potential to scale our approach to larger domains, and will make it applicable even in situations where it is difficult to obtain a set of training examples that contains a reasonable sampling of worlds that are likely to be relevant to the agent. This line of work will require the development of techniques for effectively exploring the world while learning a model, much as is done in reinforcement learning. In the longer term, we would like these online algorithms to learn not only operators and concept predicates, but also useful primitive predicates and motor actions.

## Acknowledgments

This material is based upon work supported in part by the Defense Advanced Research Projects Agency (DARPA), through the Department of the Interior, NBC, Acquisition Services Division, under Contract No. NBCHD030010; and in part by DARPA Grant No. HR0011-04-1-0012 .